\documentclass[letterpaper]{article} 
\usepackage[preprint]{aaai2027}  
\usepackage[hyphens]{url}  
\usepackage{graphicx} 
\usepackage{natbib}  
\usepackage{caption} 
\usepackage{booktabs}
\usepackage{xcolor}
\usepackage{amsmath,amssymb}
\usepackage{tikz}
\usetikzlibrary{arrows.meta,positioning,fit}

\newif\ifdraft
\draftfalse

\newcommand{\pend}[1]{\ifdraft\textcolor{blue!70!black}{#1}\else#1\fi}

\newcommand{\edward}{\textsc{Route-option harness}}

\title{Where It Breaks: Diagnosing Perception, Memory, and Decision Failures\\of a Vision-Language Agent in Pac-Man}
\author{Xinglu Wang, Edward Wang, Yong Zhang, Zhenan Fan}
\affiliations{\texttt{yanawlx@gmail.com}}

\begin{document}

\maketitle

\begin{abstract}
Spatial agents must interpret observations, maintain consistent spatial memory and make decisions over long interactions. We use Pac-Man as a testbed to study the training challenges behind these capabilities in a Qwen3.5-9B-based agent. Our research trajectory progresses from primitive-action RL on text-based mazes to visual inputs with an internal-state-assisted harness, then toward observation-driven memory and joint policy--harness improvement. The main analysis focuses on text-based maze RL; supplementary sections report visual-adaptation and harness experiments and discuss subsequent directions. Text-based maze RL requires curriculum and training-recipe design to overcome cold starts; normal-start clear rates on eight held-out mazes rise from 0\% to a peak of 81.25\%. In the second step, model adaptation and harness improvements raise the clear rate from 26.6\% to 96.9\% on one fixed map with three extra lives. These results highlight three development lessons: use curricula to address cold starts, establish visual grounding before language adaptation, and treat harness design as a determinant of agent performance. They provide practical guidance for subsequent VLM training and spatial-agent development.
\end{abstract}

\begin{figure*}[t]
\centering
\includegraphics[width=\textwidth]{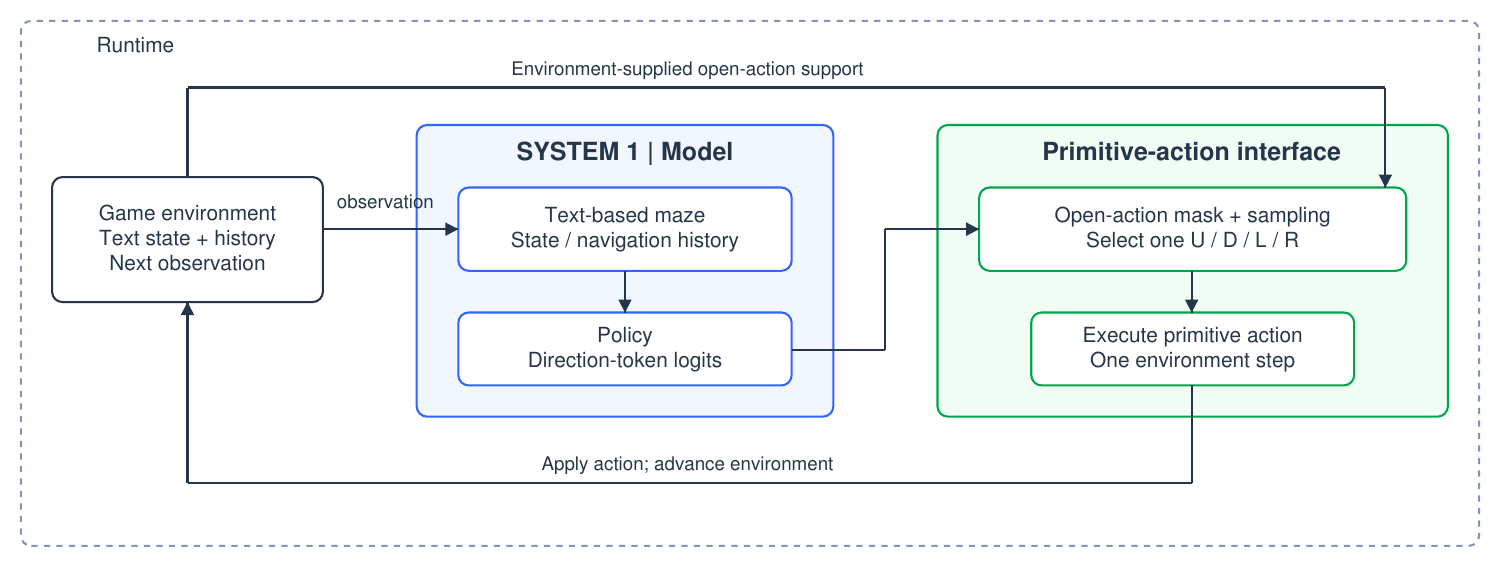}
\caption{Step 1 execution loop. The environment supplies textual observations and open-action support. Policy logits are masked and sampled into one primitive direction; the interface executes it and the environment returns the next observation.}
\label{fig:step-one-system}
\end{figure*}

\section{Introduction}
Spatial agents must interpret their surroundings from observations that may be misread or incomplete, update their memory of maps and moving objects throughout interaction, and make successive decisions toward a goal. They must incorporate new information while maintaining consistent spatial memory and considering how current actions affect later choices. Pac-Man brings these requirements together in an intuitive interactive setting: an agent collects distributed targets, tracks moving threats, and balances progress against risk.

Despite its apparent simplicity, Pac-Man results rely on different information conditions and evaluation objectives. Some systems receive game-provided maze, pellet and ghost states rather than infer the environment from screenshots, as in JevMan's direction-selection interface.\footnote{\url{https://opper.ai/jevman-benchmark/leaderboard}} HRA instead uses object-position representations extracted from images; its maximum-score experiment additionally memorizes successful action sequences that do not generalize~\citep{vanseijen2017hybrid}. Pixel-based RL has also made substantial progress~\citep{mnih2015human}: Agent57 exceeds its human-score reference on all 57 Atari games, including Ms.\ Pac-Man, but high cumulative scores do not establish reliable level completion or unseen-maze generalization~\citep{badia2020agent57}. Increasing maze scale presents an additional learning and control challenge. Information access, assistance, task scale and evaluation objectives therefore matter when interpreting performance.

Even when strong commercial models perform well in particular configurations, control through frequent model calls incurs inference costs and latency; per-call overhead accumulates over long episodes and training samples. For systems that depend on closed models whose training data, objectives and recipes are undisclosed, final performance alone also provides limited access to how capabilities were acquired and to reproducible VLM-training lessons. We therefore ask not only whether an existing model can play well, but \emph{which challenges arise when training a long-horizon spatial agent, and how development records can illuminate learning progress and failure}. Pac-Man offers inspectable trajectories and adjustable task conditions as a testbed for this investigation, with potential lessons for developing agents in other spatial tasks.

Our research trajectory has four steps. \textbf{First: primitive-action RL on text-based mazes.} The model directly outputs up, down, left or right, without a candidate-strategy harness. Removing image inputs still leaves a cold-start challenge: the initial policy has a 0\% normal-start clear rate at the final tier, motivating the design of suitable training recipes. This step builds a foundation for subsequent VLM decision training with visual inputs. \textbf{Second: visual inputs and an engine-state-assisted harness.} To move toward visual spatial agents, we introduce screenshots into the model input. This transition first raises a perception-adaptation question: game images may differ from the visual backbone's pretraining distribution, and efficiently adapting to this scenario requires its own training recipe. The first step takes approximately 175 hours of training, motivating a higher-level action interface: the harness uses internal game state to generate and check candidate strategies, the model selects among them using visual input, and the harness executes the selected strategy as an action chunk, reducing the burden of choosing each primitive direction. We study visual adaptation and control mechanisms separately, reporting their results and lessons. \textbf{Third: an observation-and-feedback-driven harness.} Screenshots and permitted public feedback support memory maintenance, with private state reserved for diagnosis and scoring. This step examines challenges from reading errors, memory updates and missing control information. \textbf{Fourth: joint RL and harness improvement.} This step combines policy learning with improvements to perception, memory and runtime assistance.

Using the development of a Qwen3.5-9B-based Pac-Man agent as a case study~\citep{qwen35}, we present an early empirical study and research perspective. The main text focuses on primitive-action RL on text-based mazes. Supplementary sections present the completed visual-adaptation and state-assisted harness experiments, preliminary observation-driven memory diagnoses, and directions for joint policy and harness improvement.

Our main results are as follows. In the first step, normal-start validation clear rates on eight held-out mazes increase from 0\% to a peak of 81.25\%. In the second step, model adaptation and harness improvements raise the clear rate from 26.6\% to 96.9\% on one fixed map with three extra lives.

These experiments provide three development lessons. \textbf{First, use curriculum design to address cold starts.} Progressing from easier tasks provides a learnable starting point for primitive-action RL. \textbf{Second, stage visual grounding and language adaptation.} We first freeze the language component while training vision, then freeze vision while adapting language, to reduce the risk of relying on training-layout shortcuts while neglecting visual inputs. \textbf{Third, harness design affects realized capability.} Candidate strategies, risk checks, and execution and fallback rules determine what the model can choose and execute, and their limitations affect final performance. As an early exploration, this study also draws on experimental and development records to identify key obstacles that may arise in moving from textual decision-making to visual spatial agents, including perception adaptation, spatial memory maintenance, and interactions between the policy and execution framework, and proposes testable directions for further research. These lessons and research perspectives inform subsequent VLM training and spatial-agent system design.

\section{Problem Definition}\label{sec:problem}
\subsection{General interaction and functional decomposition}\label{app:agent-decomposition}
A long-horizon spatial agent must reach a goal through successive observations and actions. At decision $t$, the environment has true state $s_t$ and supplies the permitted observation $o_t$ and feedback $e_t$ from the preceding interaction. The agent selects action $a_t$, advancing the environment to the next observation, and repeats until the goal or an episode limit is reached. Effective control requires interpreting the observation, maintaining relevant spatial information over time, and choosing actions that support later progress.

We decompose spatial interaction into perception, state maintenance and decision:
\begin{align}
    z_t &= V_\theta(o_t) && \text{(perception)} \\
    m_t &= U(m_{t-1}, z_t, e_t) && \text{(state maintenance)} \\
    a_t &= \pi_\phi(o_t, m_t, \mathcal{C}_t) && \text{(decision)}
\end{align}
Here $V_\theta$ interprets $o_t$ into $z_t$, such as map cells and entity positions. The update rule $U$ combines previous maintained state $m_{t-1}$, the new interpretation and permitted feedback $e_t$ into $m_t$. The policy $\pi_\phi$ selects $a_t$ from the observation and maintained information, with $\mathcal C_t$ denoting route candidates when supplied by a harness. Parameters $\theta$ and $\phi$ belong to perception and decision, respectively. The observation interface specifies which state summaries and candidates reach the policy. Errors in interpretation can persist in maintained state and affect later decisions.

\subsection{Step 1: text-based maze control}
We study this spatial interaction problem in Pac-Man under textual observations. The goal is to clear ordinary pellets while avoiding moving ghosts. A normal-start episode begins at the game's initial state and ends on completion, life exhaustion or the applicable step limit.

Here $s_t$ is the simulator state, and $o_t$ is the supplied text-based maze with live-state and navigation-history summaries, including positions and available exits. Feedback $e_t$ appears in the updated observation after each action. The interface supplies the spatial representation $z_t$ and history $m_t$ within $o_t$, serving the functions of $V_\theta$ and $U$; the learned policy $\pi_\phi$ selects $a_t\in\texttt{\{U,D,L,R\}}$ from the legal-direction support $\mathcal A_t$. The policy selects primitive directions directly; $\mathcal C_t$ denotes the route candidates used in the later visual harnesses. Figure~\ref{fig:step-one-system} shows the interaction loop; interface details appear in Appendix~\ref{app:environment-interface}.

\section{Training Method}\label{sec:system}
We post-train the Qwen3.5-9B primitive-action policy through progressively harder tasks, rehearsal and localized practice from saved states. This curriculum addresses cold start and the difficulty of learning long action sequences from episode outcomes. The restart-training configuration is summarized in Table~\ref{tab:fsr-config}.

\subsection{Curriculum, rehearsal and restart starts}
The curriculum changes maze geometry, objectives and ghost counts across seven stages (Table~\ref{tab:curriculum-progression} and Figure~\ref{fig:historical-tier-mazes}). Before the final stage, episodes begin at the game's initial state. During development, we found that small changes in restart position could make the remaining task much easier, complicating early-stage difficulty calibration. We therefore reserve failure-state restarts for the final stage. In the curriculum of Figure~\ref{fig:ascii-curve335}, a small rehearsal share is allocated only to the immediately preceding stage.

Once the policy has acquired some ability, restarts concentrate practice on difficult suffixes to improve training efficiency. Starting states are sampled from an evolving bank of simulator states saved before deaths, including deaths within games that eventually clear. Final-stage rehearsal can also use these saved states. When no suitable state is available, an episode begins normally. Restoration preserves remaining objectives and lives within the original total step budget. Sampling and restoration procedures are specified in Appendix~\ref{app:fsr-training-details}.

\subsection{Task-adapted group-based policy learning}\label{sec:policy-objective}
We train with a clipped, group-relative policy objective based on GRPO~\citep{shao2024deepseekmath}. Each group contains complete episodes sampled from a shared starting state. Episode returns include life and speed bonuses; their group mean provides the baseline, without group-standard-deviation scaling. The resulting episode-level advantage is shared across sampled actions. This supplies coarse credit for a long action sequence: a positive-advantage episode can reinforce a poor move even when later actions compensate for it. We therefore examine training dynamics alongside completion rates. Appendix~\ref{app:grpo-conventions} specifies the objective variants and implementation details.

\section{Experiments and Results}\label{sec:findings}\label{sec:evaluation}
\begin{figure*}[!t]
\centering
\includegraphics[width=\textwidth]{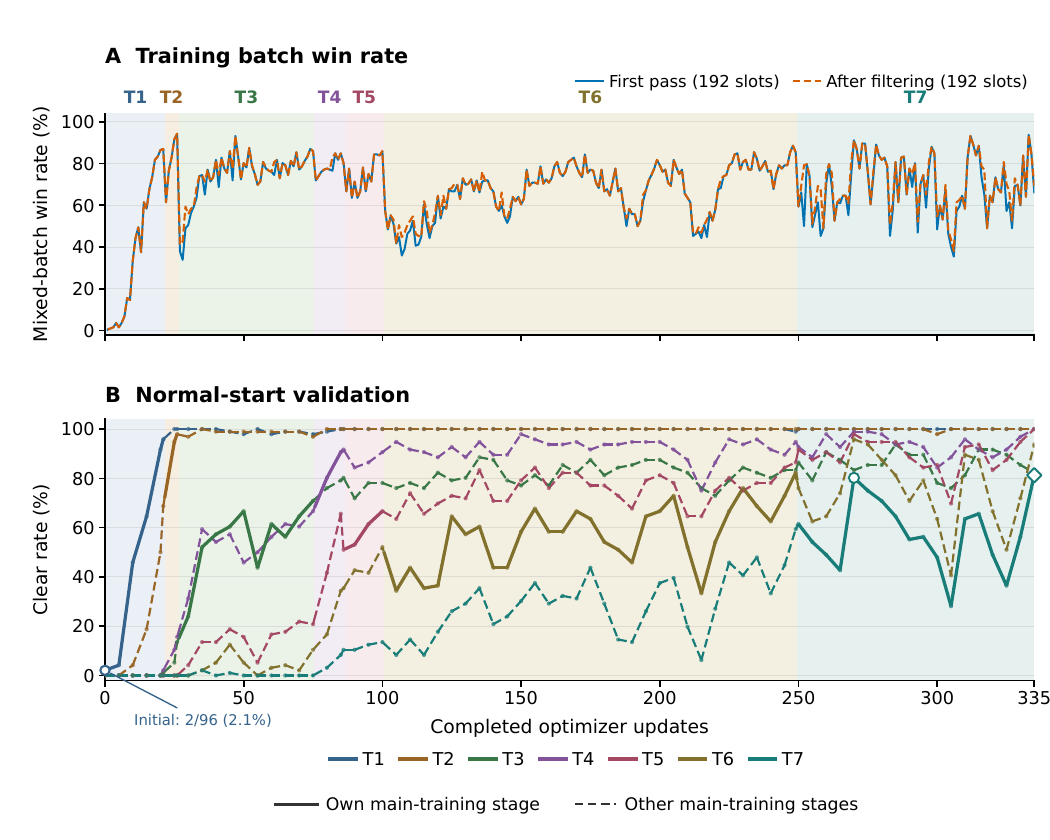}
\caption{Primitive text-based maze learning. A: mixed-start training clear rates before/after filtering (192 slots each); B: validation from normal resets. Backgrounds identify the main training tier; validation lines are solid in their own main stage and dashed otherwise, including periods with rehearsal. The circle and diamond mark validation at checkpoints 270 and 335. The horizontal axis counts optimizer updates.}
\label{fig:ascii-curve335}
\end{figure*}
\paragraph{Evaluation setup.}
Step 1 uses the textual primitive-action loop in Figure~\ref{fig:step-one-system}. Normal-start validation uses the ordinary reset, three extra lives (besides the current life) and a 512-environment-action cap, with no restored state, action prefix or outcome filtering. A clear requires collecting all ordinary pellets before life exhaustion or the cap. Each stage has eight fixed held-out layouts with twelve samples each; completion is the fraction of these 96 attempts that clear.

Initial and trained policies are evaluated on the same layouts with matched environment seeds and action-sampling settings. The validation panel supports monitoring, checkpoint selection (including 335) and reporting. Sampling and map-separation limitations are detailed in Appendix~\ref{app:result-evaluation-accounting}.

\subsection{Transfer to held-out layouts at the same difficulty}\label{sec:rl-result}
The trained policy transfers to held-out layouts at the same difficulty: optimizer checkpoint 335 clears 78/96 normal-start attempts (81.25\%) on eight held-out layouts. The final-stage pretrained baseline clears 0/96. Checkpoint 270 already clears 77/96 (80.2\%), with the pretrained and 270 policies evaluated on matched initial states and execution seeds (Appendix~\ref{app:historical-endpoints}).

Figure~\ref{fig:ascii-curve335} shows how training and normal-start validation performance evolve through the curriculum.

The policy also improves across stages beyond direct rehearsal. After main training shifts to T7 at update 250, only T6 receives a small rehearsal share. T4 and T5 receive no further direct training yet improve from 91/96 and 83/96 clears at update 249 to 96/96 each at 335. The exposure settings and validation records are documented in Appendix~\ref{app:earlier-stage-transfer}.

\subsection{Restart-training wins measure a shorter remaining task}\label{sec:training-start-results}
Restart-training success measures a different task from normal-start completion. After restarts are introduced in the final curriculum stage, training often begins close to completion, whereas validation starts from the beginning of the game.

To examine the remaining task practiced by restarts, we use a separately sampled training run, distinct from the learning curve in Figure~\ref{fig:ascii-curve335}. Its mixed-start batch 333 clears 192/192 first-pass attempts without rejecting any group. Yet 96/132 final-stage restart attempts begin with at most fifteen of the original 118 ordinary pellets remaining, including 48 with only one (Table~\ref{tab:ascii333-objective}). The eleven restart states come from two training maps and three source episodes.

\begin{table}[!ht]
\centering\small
\begin{tabular}{lrr}
\toprule
Ordinary pellets remaining & Restart states & Attempts \\
\midrule
1 & 4 & 48 \\
10--15 & 4 & 48 \\
58--60 & 3 & 36 \\
\bottomrule
\end{tabular}
\caption{Initial objective size for the eleven T7 restart states in batch 333; each state is sampled twelve times. The original tasks contain 118 ordinary pellets. Thus 96/132 T7 restart attempts begin with at most fifteen pellets remaining, including 48 with only one. The 132 attempts reuse eleven saved states and are therefore correlated.}
\label{tab:ascii333-objective}
\end{table}

Saved states turn over rapidly, but each bank draws from only two to five maps and four to twelve source episodes. State turnover therefore does not imply broad task coverage. This shorter, concentrated practice distribution explains why high training success can coexist with lower normal-start performance: efficient suffix learning trades off against full-task coverage (Appendix~\ref{app:ascii-transfer-distribution}).

\subsection{Harder-task transfer and training stability remain obstacles}\label{sec:diagnostic-limits}
Within-stage improvement does not carry over to the next task configuration: the restart-trained policies at checkpoints 270 and 310 each clear 0/96 on T8, despite their separate T7 results of 77/96 and 61/96 (Table~\ref{tab:ascii-t8-checkpoints}). T8 changes topology and maze size as well as the ordinary-pellet objective, from 118 to 192. Together, these changes define the observed task-shift boundary. Full T8 conditions and termination counts appear in Appendix~\ref{app:ascii-transfer-distribution}.

\begin{table}[!ht]
\centering\small
\begin{tabular}{lrr}
\toprule
Checkpoint & T7 clears & T8 clears \\
\midrule
270 & 77/96 & 0/96 \\
310 & 61/96 & 0/96 \\
\bottomrule
\end{tabular}
\caption{Normal-start transfer from T7 to T8. Each checkpoint is reported on the same eight layouts within each stage; T7 and T8 have different task configurations. Detailed T8 trajectory counts are in Table~\ref{tab:ascii-t8-audit}.}
\label{tab:ascii-t8-checkpoints}
\end{table}

Training success can also drop within the same T7 failure-state-restart phase: from batch 299 to 300, first-pass clears fall from 163/192 (84.9\%) to 93/192 (48.4\%), and filtered clears from 163/192 to 105/192 (54.7\%). Sparse terminal feedback and a shared episode advantage limit action-level credit assignment: equal-return groups supply no relative policy advantage, while a positive-advantage episode can reinforce mistakes later corrected. These weak or coarse signals provide a mechanism for unstable learning. Restart banks and group composition also change, so the observed drop does not isolate their respective contributions (Appendix~\ref{app:training-drop-signal}).

\section{Step 2: Visual Grounding and Assisted Control}\label{sec:visual-step2-main}
\begin{figure*}[!t]
\centering
\includegraphics[width=\textwidth]{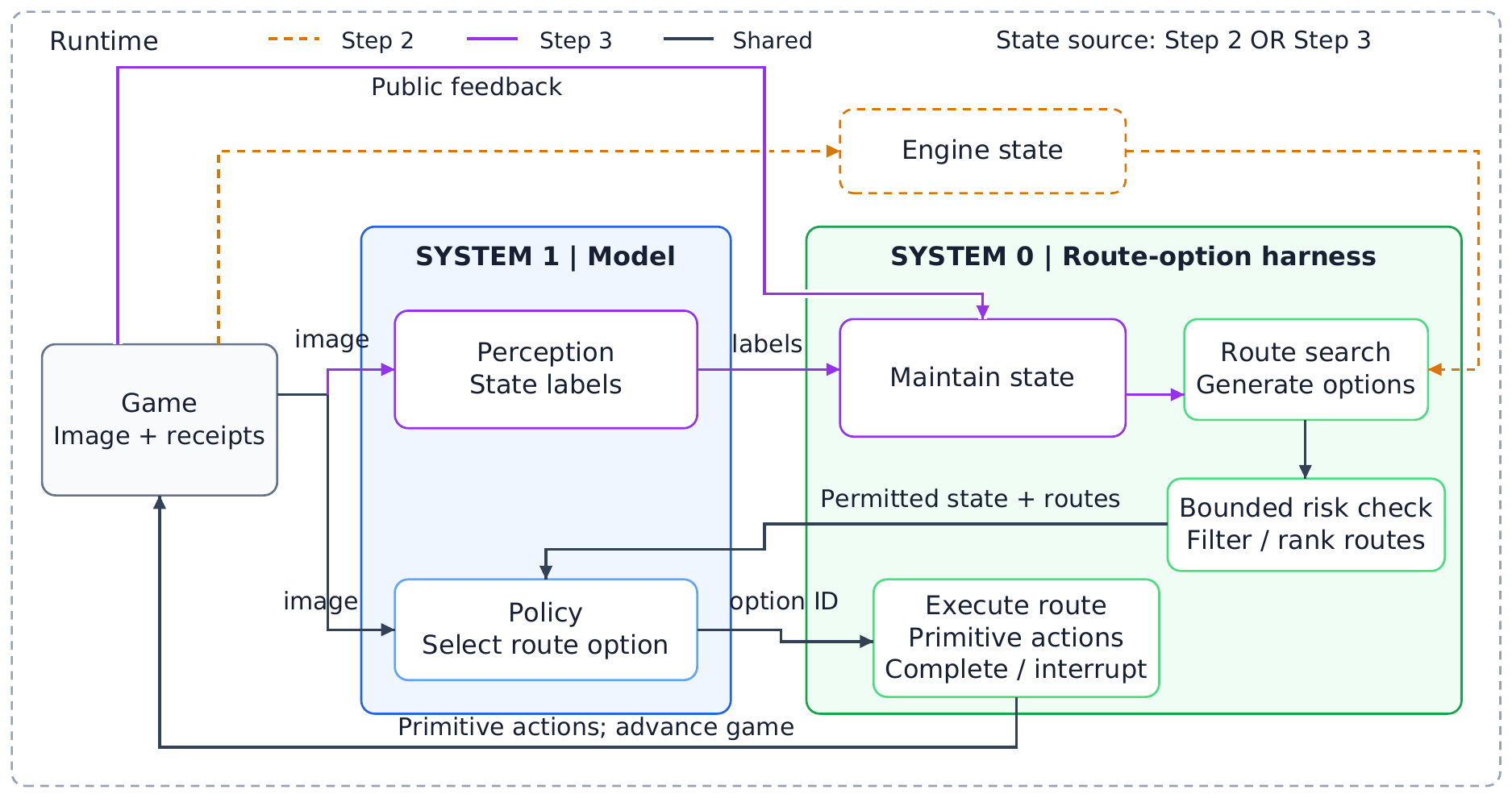}
\caption{Route-option control with alternative state sources. Step 2 uses engine state for candidate generation and bounded risk checking; Step 3 maintains state from screenshots and public feedback. The policy receives screenshots and permitted state/route summaries, selects an option, and the harness executes primitive moves until completion or interruption. Each instance enables one state source.}
\label{fig:overview}\label{fig:overview-full}
\end{figure*}

Step 2 adds visual observations and temporally extended actions to the primitive-action task. The model selects a state-dependent \textsc{Collect}, \textsc{Avoid} or \textsc{Eliminate} route, which the harness executes as primitive moves. Engine positions and ghost states support candidate generation and risk checks; the model receives screenshots, live-state/history summaries and candidate routes. Figure~\ref{fig:overview} contrasts this engine-assisted pathway with the screenshot-and-public-feedback pathway considered for Step 3. We evaluate visual adaptation and fixed-weight control separately (Appendices~\ref{sec:step-two} and~\ref{app:factorial-protocol}).

\begin{figure*}[t]
\centering
\includegraphics[width=0.98\textwidth]{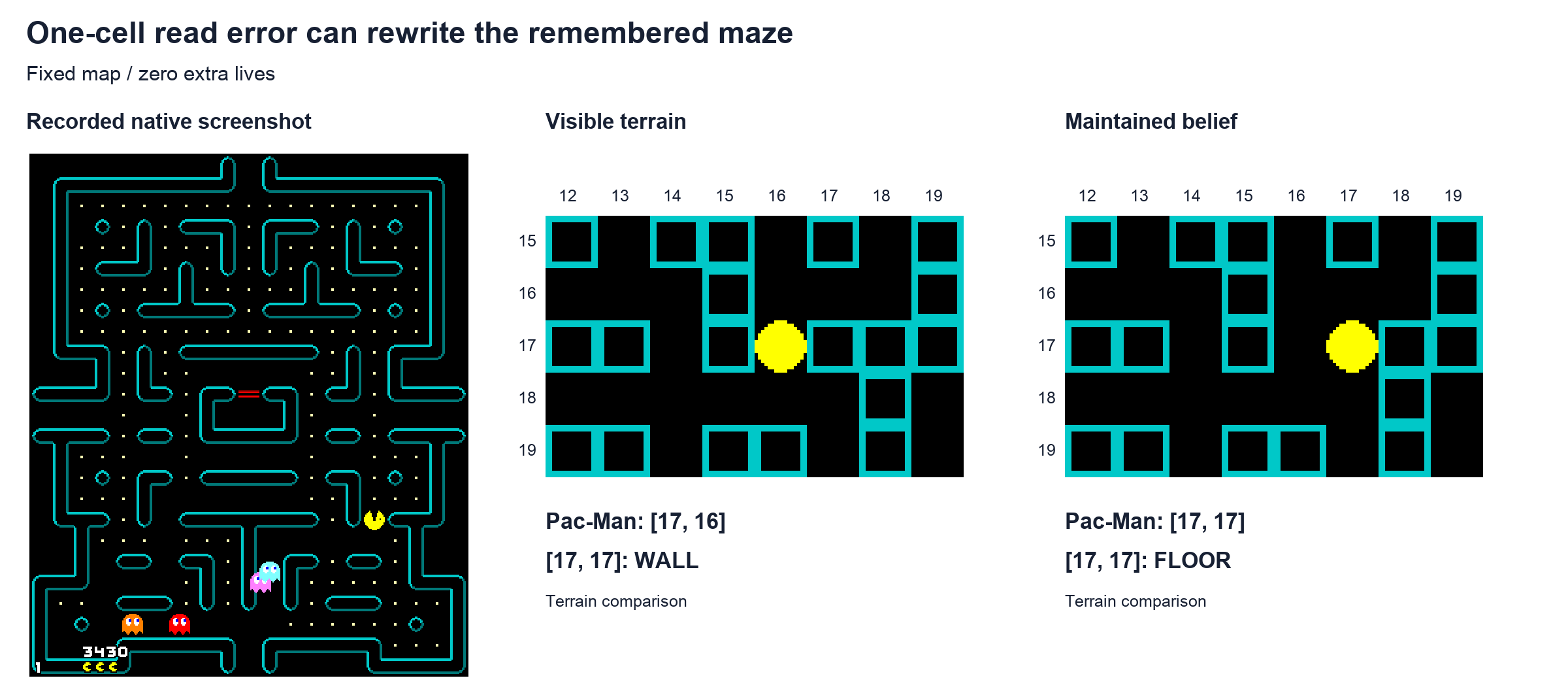}
\caption{A screenshot-feedback memory error on one fixed map. Left: the observed game image. Right: matching $5\times 8$ crops of visible labels and maintained terrain. A fallback map reading shifts Pac-Man from $[17,16]$ to $[17,17]$, causing a wall to be remembered as floor. A ghost also overlaps a wall in this scene, an unresolved engine anomaly; the example illustrates error propagation rather than isolating its full-game cause. Private state is used only to score the diagnosis.}
\label{fig:memory-audit}
\end{figure*}

\begin{table}[!ht]
\centering\small
\setlength{\tabcolsep}{3pt}
\begin{tabular}{llrr}
\toprule
\multicolumn{4}{l}{\textbf{A: Whole-map transcription}} \\
\multicolumn{2}{l}{Checkpoint} & Exact & Accuracy \\
\midrule
\multicolumn{2}{l}{Single-size vision} & 256/531 & 48.2\% \\
\multicolumn{2}{l}{Multi-size vision} & 401/531 & 75.5\% \\
\multicolumn{2}{l}{Multi-size vision + language LoRA} & 427/531 & 80.4\% \\
\midrule
\multicolumn{4}{l}{\textbf{B: Fixed adapted model: normal-start clears}} \\
Execution & Gate & 0 extra & 3 extra \\
\midrule
Target replan & basic & 7/64  & 40/64 \\
Target replan & improved   & 18/64 & 59/64 \\
Bound route & basic & 9/64  & 39/64 \\
Bound route & improved   & \textbf{27/64} & \textbf{62/64} \\
\bottomrule
\end{tabular}
\caption{Step 2 separates screenshot grounding from assisted control. A uses the same 531-item schema-constrained screenshot stream. B holds adapted-model weights, candidates and fallback fixed on one map, with 64 paired seeds per arm and separate extra-life budgets. Full protocols remain in Appendices~\ref{sec:step-two} and~\ref{app:factorial-protocol}.}
\label{tab:step2-main-results}\label{tab:visual-adaptation-stream}\label{tab:factorial}
\end{table}

\subsection{Visual adaptation improves whole-map transcription}
On the same 531-item level-1 screenshot stream, schema-constrained whole-map transcription improves from 256/531 (48.2\%) for single-size visual adaptation to 401/531 (75.5\%) for multi-size visual adaptation, then 427/531 (80.4\%) with language-only LoRA (Table~\ref{tab:visual-adaptation-stream}A). Each observation is a screenshot-stream item; game completion is measured separately below. These checkpoint results support establishing visual grounding before adapting the response interface.

\paragraph{Staged visual adaptation.}
The first stage targets the visual representation: with the language component frozen, the vision encoder and multimodal merger learn to transcribe visible terrain and entities. Labels mark occluded cells as unknown. The second stage holds this representation fixed and trains language-only LoRA on visual questions, adapting the answer interface while retaining the learned visual features. This design assigns visual grounding to the encoder and merger, and response adaptation to the language component. Preprocessing, output formats and reproduction details are specified in Appendix~\ref{app:d5a-reconstruction}.

\paragraph{Transfer of the visual reader.}
The supporting transcription evaluations also vary layout family and maze size. The multi-size visual models without and with language LoRA obtain 50/64 and 51/64 exact maps on an unseen family, and 182/216 and 189/216 on unseen sizes (Appendix~\ref{sec:step-two}). These measurements extend the reader evaluation beyond the single screenshot stream. The smaller gains from language LoRA on these sets also show why improvement on one stream should be checked against a spatially varied evaluation. They assess exact structured readings, whereas the control experiment below holds the game map fixed. Consequently, spatial coverage of the reader and robustness of the controller remain complementary dimensions of evaluation.

\paragraph{Grounding and control measure different capabilities.}
Whole-map exactness requires every represented cell to match, so a single error can invalidate an otherwise accurate reading. Game completion instead measures a sequence of decisions supported here by engine-derived summaries and checked routes. The screenshot stream evaluates the adapted reader; the fixed-weight factorial evaluates how control mechanisms affect the same learned model.

\subsection{Fixed weights expose the harness contribution}
With adapted-model weights, candidate generation and risk-ranked fallback fixed, jointly improving route gates and binding execution raises completion from 40/64 to 62/64 (62.5\% to 96.9\%) with three extra lives; with zero extra lives, the change is 7/64 to 27/64 (Table~\ref{tab:factorial}B). Each arm uses 64 paired execution seeds on one $25\times21$ map, four normal ghosts, no fruit, temperature 0.7 and a 512-native-step limit. These panels measure control in the engine-state-assisted, fixed-map setting.

\paragraph{From a route choice to executed actions.}
The route menu changes with the state, so its code is not a fixed direction command. \textsc{Collect} proposes pellet targets with an escape route, \textsc{Avoid} seeks a junction with enough exits, and \textsc{Eliminate} pursues an edible ghost when time permits. Candidate generation therefore supplies part of the spatial planning structure before the model chooses. The harness then turns the selected target or route into primitive actions and checks whether execution can continue. All factorial arms share this proposal scaffold, ordering and fallback, so the comparison changes commitment and gate rules within that scaffold.

Improved risk gates combine a two-move motion-envelope check with near-target escape protection. Target replanning recomputes the route after each move; bound execution follows it until completion, invalidation or safety interruption.

\paragraph{Why the gate and commitment interact.}
A route can lead toward a valid target yet leave the player poorly placed after reaching it. The improved gate therefore checks both short-horizon motion and the possibility of escaping a nearby target: for a \textsc{Collect} target within two moves, it searches up to four additional moves toward an escape junction. This makes route eligibility depend on what comes immediately after the apparent local success. Binding then preserves the selected path across several environment transitions, while continued safety checks can interrupt it. The two mechanisms address different failure modes: admitting a hazardous route and changing or repeating decisions while a viable route is already being executed.

\paragraph{The execution rule must fit the policy.}
Under improved checks, binding changes the unadapted model from 11/64 to 9/64 clears with zero extra lives and leaves 37/64 unchanged with three, while the adapted model improves to 27/64 with zero extra lives and 62/64 with three. Execution rules therefore need to fit the learned route selector.

\paragraph{Execution cost on matched successes.}
Under improved gates, bound execution saves an average of 47.88 model calls with zero extra lives and 26.91 with three, measured on the 8 and 57 paired games that both strategies complete. These savings measure decision frequency on jointly successful games, rather than average cost across all attempts (Appendix~\ref{app:factorial-protocol}).

The broader system progression combines model adaptation and harness changes: under the three-extra-life condition, base Qwen with target replanning and basic gates clears 17/64 (26.6\%), while the adapted model with bound routes and improved gates clears 62/64 (96.9\%) (Table~\ref{tab:factorial-complete}).

Together, the transcription and fixed-weight control results show why visual grounding and realized task capability need separate measurements. Candidate eligibility and execution commitment change which decisions the model can make and how those decisions unfold, even with its learned weights fixed.

\section{Memory and Motion as Control Bottlenecks}\label{sec:memory-motion-main}
Moving from engine assistance to screenshot-derived state exposes two distinct requirements: preserving spatial structure over time and supplying motion information needed by the planner. The following case and local diagnostic examine these requirements in preliminary public-feedback systems.

\paragraph{A transient reading error can persist in memory.}
A fallback map reading places Pac-Man at $[17,17]$ instead of $[17,16]$. The state updater treats the estimated player cell as traversable and changes a correctly stored wall into floor (Figure~\ref{fig:memory-audit}). The next player reading is correct, but the terrain error remains. Thus correcting a current observation does not necessarily repair the spatial memory used by subsequent route construction. Separating static terrain, pellet events and dynamic actors prevents actor estimates from directly rewriting walls. A ghost also overlaps a wall in this scene, an unresolved engine anomaly; this single case identifies error propagation rather than its frequency or effect on full-game completion (Appendix~\ref{app:historical-memory-trace}).

\paragraph{Correct positions do not supply all motion information.}
A separate diagnostic uses 431 pre-action states from four full games in the later public-feedback system. Player positions and pellet sets are correct, yet 291 states contain at least one active ghost with unknown heading. Across all 431 recorded states, the gate rejects every tested action in 292 states. Replacing uniquely matched ghost headings with private truth reduces this count to 92/431; replacing only subcell pixel positions leaves it at 292/431 (Table~\ref{tab:all-state-gate}). The planner therefore needs motion information beyond correct grid locations. These are local input substitutions on correlated states, with private truth used only for diagnosis; they do not measure new trajectories or a completion-rate improvement. Together, the two analyses motivate protecting stable map structure while maintaining the dynamic information required for route checks.

\section{Discussion}
\paragraph{Learnability and task coverage.}
A curriculum must offer attainable early successes while increasing decision difficulty. Restart practice targets difficult suffixes, but concentrated starting states can make training completion a poor measure of full-game ability. Normal-start validation complements this practice by testing whether the policy can reach those suffixes from the intended initial condition.

\paragraph{Robustness and attribution.}
Repeated training seeds, complete map-split information and an independent reporting set would strengthen robustness claims. Matched comparisons can separate curriculum, restart and reward-design effects under a common observation interface.

\paragraph{Toward visual spatial agents.}
Three questions follow. Can staged visual adaptation improve full games on unseen layouts, rather than exploit familiar geometry? Can typed state updates and deployable heading estimates prevent persistent memory errors? How should policy learning account for candidate coverage and execution constraints? These require full-game comparisons alongside perception and state accuracy, with information access and action budgets fixed (Appendices~\ref{app:perception-panels},~\ref{app:public-memory-followup} and~\ref{expanded:sec:discussion}).

\section{Conclusion}
The Pac-Man training trajectory attains substantial normal-start improvement under textual observations. Restart-objective and source-coverage analyses show how remaining-task difficulty and source diversity shape training wins relative to full-game validation. Step 2 improves screenshot grounding and shows that risk checking and route execution change completion rates even with fixed model weights. Preliminary diagnostics expose persistent map corruption and missing motion information, supporting joint evaluation of training starts, perception, memory and control.

\clearpage
\bibliography{references}

\clearpage
\appendix
\section{Shared Environment, Interfaces and Evaluation}
\label{app:expanded-methods}\label{app:development-records}\label{app:experiment-protocols}\label{app:audited-details}\label{app:interface-detail}
\subsection{Broader development trajectory}
Our research trajectory has four steps. \textbf{Step 1: primitive-action RL on text-based mazes.} The model directly outputs up, down, left or right, without a candidate-strategy harness. Removing image inputs still leaves a cold-start challenge: the initial policy has a 0\% normal-start clear rate at the final tier, motivating the design of suitable training recipes. This step builds a foundation for subsequent VLM decision training with visual inputs. \textbf{Step 2: visual inputs and an engine-state-assisted harness.} To move toward visual spatial agents, we introduce screenshots into the model input. This transition first raises a perception-adaptation question: game images may differ from the visual backbone's pretraining distribution, and efficiently adapting to this scenario requires its own training recipe. The primitive-action interface motivates exploring a higher-level action interface: the harness uses internal game state to generate and check candidate strategies, the model selects among them using visual input, and the harness executes the selected strategy as an action chunk, reducing the burden of choosing each primitive direction. We study visual adaptation and control mechanisms separately, reporting their results and lessons. \textbf{Step 3: an observation-and-feedback-driven harness.} Screenshots and permitted public feedback support memory maintenance, with private state reserved for diagnosis and scoring. This step examines challenges from reading errors, memory updates and missing control information. \textbf{Step 4: joint RL and harness improvement.} This step combines policy learning with improvements to perception, memory and runtime assistance.

\subsection{Step 1 curriculum catalog}\label{app:curriculum-catalog}
\begin{figure*}[!t]
\centering
\includegraphics[width=\textwidth]{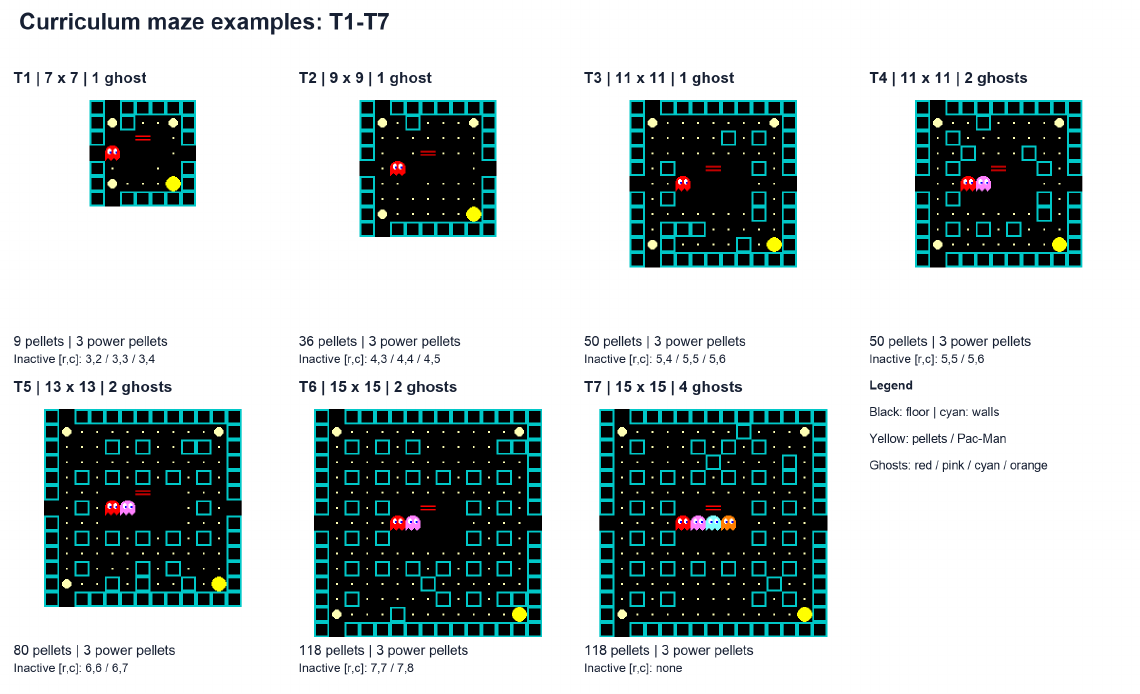}
\caption{Validation maze examples for T1--T7. Grid dimensions include walls; active ghost counts and inactive spawn coordinates are shown. The yellow player glyph marks the normal start without indicating heading. Pellets, power pellets, doors and portal openings follow the map cells. Each example uses three extra lives and a 512-action budget.}
\label{fig:historical-tier-mazes}

\end{figure*}

Table~\ref{tab:curriculum-progression} summarizes the stage geometry and objectives; Figure~\ref{fig:historical-tier-mazes} shows one normal-start example per stage. The examples illustrate stage geometry rather than every training start.

\begin{table}[!ht]
\centering
\begingroup\small
\begin{tabular}{lrrr}
\toprule
Stage & Grid size & Ghosts & Ordinary pellets \\
\midrule
T1 & $7\times 7$ & 1 & 9 \\
T2 & $9\times 9$ & 1 & 36 \\
T3 & $11\times 11$ & 1 & 50 \\
T4 & $11\times 11$ & 2 & 50 \\
T5 & $13\times 13$ & 2 & 80 \\
T6 & $15\times 15$ & 2 & 118 \\
T7 & $15\times 15$ & 4 & 118 \\
\bottomrule
\end{tabular}
\endgroup
\caption{Curriculum progression, illustrated by one validation layout at each stage. Grid sizes include walls and are rows $\times$ columns. Layouts also change across stages; the ghost-count transitions are not isolated interventions. Full grids appear in Figure~\ref{fig:historical-tier-mazes}.}
\label{tab:curriculum-progression}
\end{table}

\subsection{Problem Setup}\label{expanded:sec:problem}

\paragraph{Environment scope.}
Step 1 uses the textual task in Section~\ref{sec:problem}. The visual panels use grid-based Pac-Man mazes up to $25\times 21$ cells, four ghosts and a 512-environment-action horizon. Each comparison specifies either zero or three extra lives. The native rules below apply to the paired engine/public-feedback and action-head experiments; the earlier visual curriculum has its own task settings.

\paragraph{Environment interface.}\label{app:environment-interface}
Resetting the environment with a seed or a saved game state produces an RGB image and game information. A primitive action updates the game and returns the next image, score increment and termination status. Learning rewards are computed separately from game scores. The policy's information depends on the observation condition in Table~\ref{tab:interface-routing}; route choices are translated into primitive moves by the harness.

\begin{table*}[t]
\centering
\begin{tabular}{p{0.19\textwidth}p{0.72\textwidth}}
\toprule
Information & Use in the agent system \\
\midrule
Textual observation & Step 1 receives the text-based maze, live-state summaries and navigation history, and selects a legal direction. \\
Screenshot & The visual reader and visual policy receive the current image. Hidden or occluded cells remain uncertain. \\
Engine-derived state & The engine-assisted harness uses positions and ghost states for route generation and risk checks. The policy receives selected state/history and route summaries. \\
Public feedback & The later screenshot-feedback system receives elapsed time, relative player movement, localized pickups, deaths, respawns and terminal events. \\
Maintained state & Screenshot-feedback memory combines visible terrain, pellet changes and actor estimates; private engine values do not repair its inputs. \\
Private game state & Hidden maps, exact coordinates, ghost internals and random states are reserved for evaluation and controlled diagnostic comparisons in the public-feedback condition. \\
Score and reward & Game score, learning reward and binary completion are measured separately. \\
\bottomrule
\end{tabular}
\caption{Information access in the textual, engine-assisted and screenshot-feedback systems. The public-feedback variants differ in their event detail; these differences are stated with their experiments.}
\label{tab:interface-routing}
\end{table*}

\paragraph{Primitive actions.}
The environment supports movement and staying still, while Step 1 outputs only the four direction tokens. Each action advances the game before the next observation. The legal-action mask restricts passability; moving threats and future escape routes remain policy decisions. The available Step 1 input description specifies state and navigation summaries but not every individual ghost or timer field.

\paragraph{Observation conditions.}
\emph{Engine assistance} supplies game-derived fields to the harness and selected summaries to the model.
\emph{Screenshot feedback} builds state from images, public interaction events and known game rules, including portals and respawn. The later variant reports realized relative displacement, a richer signal than moved/blocked feedback. Absolute engine coordinates and ghost targets, timers and paths are restricted to evaluation.
\emph{Text-based maze control} supplies text and state summaries instead of image transcription. External screenshot-plus-structured-state baselines retain their additional state inputs.

\paragraph{Metrics.}
Besides completion, we report pellet clearance, deaths per episode, timeouts, harness aborts (kept separate from deaths), model calls, and environment steps. Static perception is measured independently by whole-map exact match and per-entity accuracy. Generalisation is reported separately for new random seeds on seen layouts and for held-out layouts. Each panel retains its own layout/seed split and denominator rather than a universal train/test pool. A new execution seed on one map is not a held-out layout. The action-head figure shows marginal Wilson 95\% intervals; paired outcomes and matched-success costs are reported separately. The perception panels distinguish frames, repeated decodes and query records (Appendix~\ref{app:perception-panels}).

Appendix~\ref{app:paired-uncertainty} distinguishes marginal rates, paired differences and clustered/repeated observations; its intervals refer to their explicit fixed-map sampling model.

\subsection{System and Diagnostic Method}\label{expanded:sec:system}

Figure~\ref{fig:overview} in Section~\ref{sec:visual-step2-main} shows the two state sources and their shared execution loop.

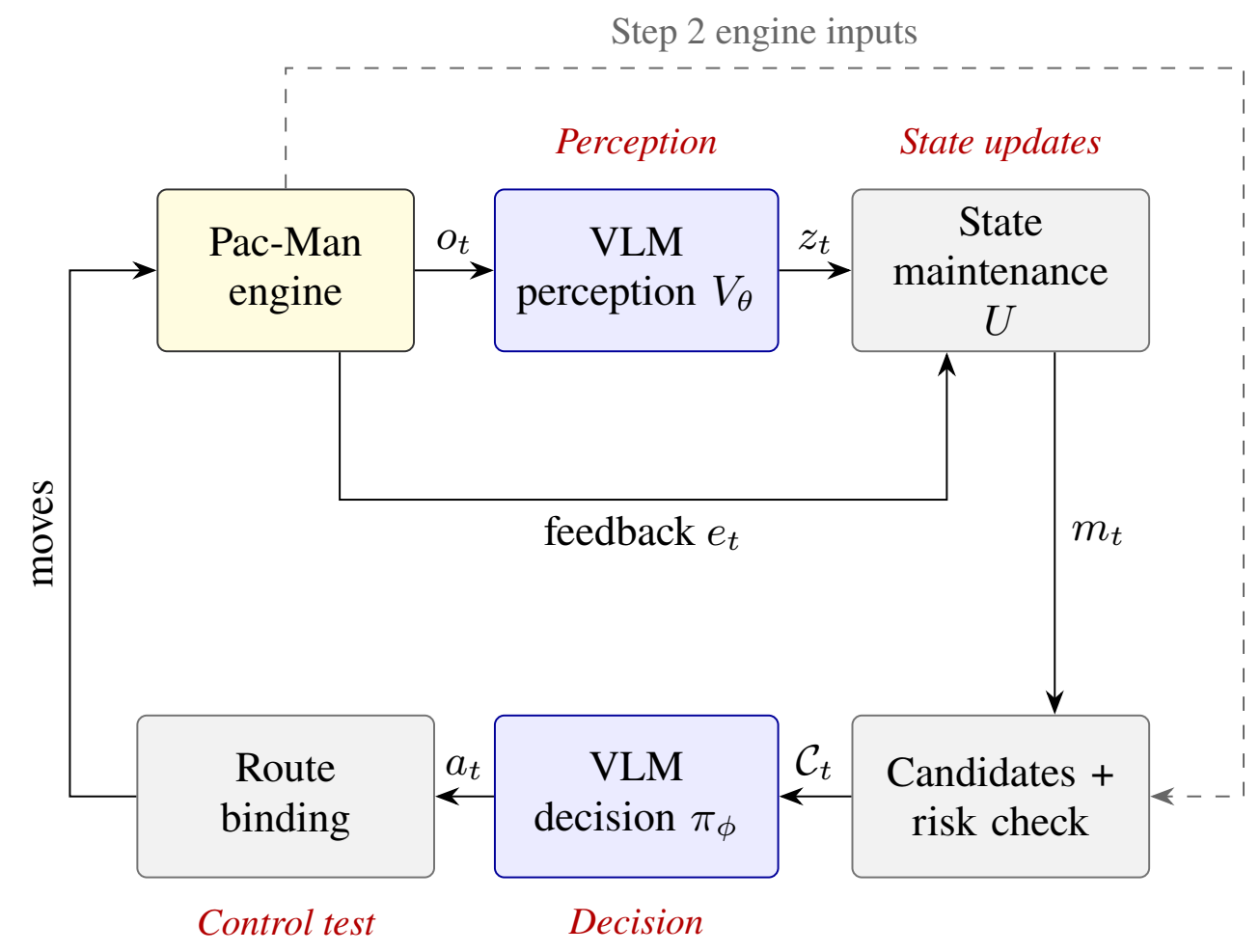
\begin{figure}[t]
\centering
\resizebox{\columnwidth}{!}{%
\begin{tikzpicture}[
 font=\fontsize{9}{10.5}\selectfont, >={Stealth[length=1.8mm]},
 learn/.style={draw=blue!60!black,fill=blue!8,rounded corners=2pt,minimum height=12mm,text width=19mm,align=center,inner sep=1mm},
 code/.style={draw=black!55,fill=black!5,rounded corners=2pt,minimum height=12mm,text width=20mm,align=center,inner sep=1mm},
 env/.style={draw=black!80,fill=yellow!15,rounded corners=2pt,minimum height=12mm,text width=17mm,align=center,inner sep=1mm},
 tag/.style={font=\fontsize{8}{9}\selectfont\itshape,text=red!70!black}
]
\node[env] (env) at (0,0) {Pac-Man\\engine};
\node[learn] (perc) at (26mm,0) {VLM\\perception $V_\theta$};
\node[code] (mem) at (53mm,0) {State\\maintenance\\$U$};
\node[code] (cand) at (53mm,-39mm) {Candidates +\\risk check};
\node[learn] (dec) at (26mm,-39mm) {VLM\\decision $\pi_\phi$};
\node[code] (exec) at (0,-39mm) {Route\\binding};
\draw[->] (env.east) -- node[above]{$o_t$} (perc.west);
\draw[->] (perc.east) -- node[above]{$z_t$} (mem.west);
\draw[->] ([xshift=4mm]mem.south) -- node[right]{$m_t$} ([xshift=4mm]cand.north);
\draw[->] (cand.west) -- node[above]{$\mathcal{C}_t$} (dec.east);
\draw[->] (dec.west) -- node[above]{$a_t$} (exec.east);
\draw[->] (exec.west) -- (-16mm,-39mm) -- node[left,rotate=90,anchor=south]{moves} (-16mm,0) -- (env.west);
\draw[->] ([xshift=4mm]env.south) -- (4mm,-17mm) -- node[below]{feedback $e_t$} (49mm,-17mm) -- ([xshift=-4mm]mem.south);
\draw[->,dashed,black!60] (env.north) -- (0,15mm) -- node[above,font=\fontsize{8}{9}\selectfont]{Step 2 engine inputs} (71mm,15mm) -- (71mm,-39mm) -- (cand.east);
\node[tag,above=0.8mm of perc] {Perception};
\node[tag,above=0.8mm of mem] {State updates};
\node[tag,below=1mm of dec] {Decision};
\node[tag,below=1mm of exec] {Control test};
\end{tikzpicture}}
\caption{Route-selection pipeline for Steps 2 and 3 and diagnostic sites. Blue nodes denote learned functions; grey nodes denote harness code. Permitted feedback $e_t$ updates state separately from moves. The dashed branch provides engine-derived candidate inputs in Step 2; Step 3 uses screenshots and public feedback. Scoring paths are omitted.}
\label{fig:pipeline}
\end{figure}

Figure~\ref{fig:pipeline} separates perception, maintained state and action selection. The experiments vary both observation conditions and control mechanisms, so comparisons use the matched conditions specified for each panel.

\paragraph{Component diagnostics.}
A diagnostic intervention changes perception $z_t$, memory $m_t$ or action $a_t$ at a shared state and measures the resulting behavior. Private state may provide a diagnostic alternative but does not repair deployed inputs. Fixed-action comparisons measure state-update accuracy; new closed-loop trajectories are needed to measure game-level benefit.

We examine a memory-write error, short-window state accuracy, paired whole-harness outcomes and a small action-assistance screen. These answer different questions: local correction identifies a failure mechanism in that case, while full-episode comparisons test whether the correction improves completion.

\subsection{Game Rules and Time Accounting}\label{app:native-rules}
The following rules describe the native Pac-Man environment used in the paired engine/public-feedback and action-head panels. The implementation has its own dynamics rather than reproducing the arcade or Atari game exactly.

\paragraph{Clear, lives and collectibles.}
A level clears when the ordinary-pellet counter reaches zero and emits the native clear event. Power pellets are not required for this terminal condition. In this environment, ordinary pellets award ten raw score points and power pellets 100; this raw game score is distinct from an experiment's learning reward. Power pickup sets a 360-active-logic-tick vulnerability timer, without consuming a tick immediately on the contact frame. The public-feedback policy tracks a rule-derived interval rather than reading the hidden timer. Fruit is disabled in these panels. 0-extra-life episodes stop at the first death; the 3-extra-life setting permits three spares in addition to the current life. The episode ends at the first death under the zero-extra-life setting.

\paragraph{Native ghost and portal behavior.}
Normal ghosts initially receive paths to randomly selected ordinary-pellet cells. When a path finishes, native pathfinding targets Pac-Man; this is not an asserted recreation of four arcade ghost personalities. In this renderer, normal-ghost pupils point according to Pac-Man's relative position rather than the ghost velocity, so eye direction is not a direct velocity observation. Normal ghosts outside the house cannot re-enter through its door. Eaten ghosts return as eyes to the pen and then leave again. The horizontal/vertical portal handler relocates Pac-Man between paired map tiles, with a direction-dependent exit offset. That player handler does not establish that ghosts share the same teleport rule. Safety predictions are conditional on the observed/declared graph and bounded motion model; an unknown portal destination is not a certified safe endpoint.

\paragraph{Actions, logic ticks and model calls.}
The wrapper's nominal movement budget is sixteen logic frames per environment action, not sixteen model decisions. Native terminal states can end an action early; death/respawn handling can instead continue until the corresponding transition completes, so the recorded atomic-frame count is authoritative. A 512-step episode cap counts environment actions. A bound option can execute several such actions after one Agent selection, while a visual-feedback action may additionally require a perception query. Logic ticks, environment steps, Agent calls, perception calls and wall-clock latency are therefore distinct quantities. Model inference pauses game advancement. No result in this paper is a wall-clock interactive benchmark.

\subsection{Failure-state-restart configuration}\label{app:fsr-training-details}
Table~\ref{tab:fsr-config} gives the text-based maze restart-training settings. These settings apply to this training recipe, separately from visual-policy and action-head training.

The restart fraction is a collection request, not the realized start-state composition; fallback can change the mix. Restart initialization and the original total-budget rule are described in Section~\ref{sec:system}.

\begin{table*}[!t]
\centering\small
\begin{tabular}{p{0.23\textwidth}p{0.70\textwidth}}
\toprule
Setting & Recorded value \\
\midrule
Observation and output & Space-separated text-based maze; one response token; prompt cap 2,048 tokens. \\
Sampling and episode & Temperature 0.7; cap 512 native steps; 3 extra lives; visual module frozen. \\
Rollout and batches & 16 groups of 12 samples (192 episodes); actor microbatch 2; global batch 192. \\
Parallelism & 8 actor GPUs; tensor/pipeline/context parallel sizes all 1. \\
Adam & Constant learning rate $5\times 10^{-7}$; betas $(0.9,0.999)$; weight decay 0; gradient clipping 1. \\
PPO and ClipCov & Both PPO clipping bounds 0.05; ClipCov enabled, bounds 1--5, ratio 0.002. \\
Regularization and rewards & Explicit KL and entropy coefficients 0; success life/speed bonus coefficients 0.1/0.05. \\
Precision and probabilities & BF16; FP32 gradient accumulation and softmax; rollout log probabilities consumed directly. \\
Rollout backend & SGLang radix/prefix caching and CUDA graphs disabled. \\
Curriculum allocation & 13 current-tier groups and 3 lower-tier replay groups; earlier-stage requested slots 18.75\%. \\
Restart request & 80\% within each class; normal-start fallback if no suitable state. \\
Restart bank & Full simulator states saved before recorded deaths; lookback bins 4--7,8--15,16--32; source episodes may eventually win; evolving bank. \\
\bottomrule
\end{tabular}
\caption{Failure-state-restart training configuration. Values retain their original protocol scope; requested restart fractions allow normal-start fallback.}
\label{tab:fsr-config}
\end{table*}

\paragraph{Unspecified settings.}
Adam epsilon and the reward-filter acceptance thresholds and retry limit are unavailable. Top-$p$ 1.0 is specified for normal-start checkpoint evaluation, not for training. The objective conventions below describe the available method specification; the complete optimization settings of every historical run are unavailable.

\subsection{Primitive-action sampling}
For text-based maze restart training, let $x_t$ be the text-based maze and declared state/history at decision $t$, and let $\mathcal A_t\subseteq\{\texttt{U},\texttt{D},\texttt{L},\texttt{R}\}$ be its nonempty advertised open-action support. With policy parameters $\theta$, direction-token logits $\ell_\theta(a;x_t)$ and sampling temperature $\tau=0.7$, the supported action distribution is
\[
\pi_\theta^{\mathrm{open}}(a\mid x_t)=
\frac{\mathbf 1\{a\in\mathcal A_t\}\exp(\ell_\theta(a;x_t)/\tau)}
{\sum_{b\in\mathcal A_t}\exp(\ell_\theta(b;x_t)/\tau)}.
\]
The policy samples the executed direction $a_t$ from this distribution. Here $a$ is a direction token, $b$ indexes the supported directions and $\mathbf 1$ is an indicator. Masking makes directions outside the declared support have zero probability. The mask constrains passability; avoiding ghosts and preserving future routes remain policy decisions.

A worked example illustrates why grouping matters:
\paragraph{Episode-level grouping.} Rollouts from the same start state are compared by their episode returns. Pooling intermediate returns from different decisions into one group is misleading: the pooled mean estimates the value of no particular state, and a single death penalty inflates the standard deviation and compresses all other advantages. In a three-episode example, changing one late death from $-100$ to $+2$ flipped the advantage of an \emph{unchanged} early decision from $+0.17$ to $-1.26$ under pooling, while episode-level grouping changed it only from $-1.41$ to $-1.31$. This worked example illustrates the effect of grouping on the arithmetic.

A return-based objective can reinforce an episode that recovers from a poor move, without assigning that move a separate outcome. This motivates checking group composition and action-level evidence when learning stalls.

\subsection{Policy objective, advantages and reduction}\label{app:grpo-conventions}
\paragraph{Clipped objective.}
For textual input $x_{it}$, action $a_{it}$, episode advantage $A_i$ and clipping parameter $\epsilon$, the standard reference objective is
\begin{align}
 r_{it}(\theta)&=\frac{\pi_\theta(a_{it}\mid x_{it})}{\pi_{\mathrm{old}}(a_{it}\mid x_{it})},\\
 J_{\mathrm{clip}}(\theta)&=\mathbb{E}_{i,t}\!\left[\min\!\left(r_{it}A_i,\operatorname{clip}(r_{it},1-\epsilon,1+\epsilon)A_i\right)\right].
\end{align}
Here $\theta$ denotes the action policy, separately from perception parameters in Section~\ref{sec:problem}. The generating policy supplies the reference denominator. The implementation described below separates the PPO ratio from behavior-policy correction; its complete correspondence to every historical training run is unknown.

\paragraph{Supported probabilities.}
Both sampling and policy evaluation restrict probabilities to the open-action set and apply temperature 0.7 once. This keeps rollout and learning probabilities on the same action support.

Let $\ell^{\rm beh}_{it}$ be the behavior log probability, $\ell_{it}$ the current actor log probability and $\ell^{\rm old}_{it}=\operatorname{stopgrad}(\ell_{it})$. With one optimization step per collected batch, the two ratios are
\[
 s_{it}=\exp(\ell_{it}-\ell^{\rm old}_{it}),\qquad
 w_{it}=\min\{\exp(\ell^{\rm old}_{it}-\ell^{\rm beh}_{it}),2\}.
\]
The PPO ratio $s_{it}$ equals one at the forward pass while retaining its derivative. The detached correction $w_{it}$ compensates behavior-policy differences and is capped at 2. Its truncation means that the product need not equal the actor/behavior ratio. The experiment specifies PPO clipping 0.05; reference-policy KL is a separate term.

\paragraph{Episode advantages.}
Twelve complete episodes share an initial state. Their returns are centered within the group:
\[
 A_i=R_i-\frac1{12}\sum_{j\in g(i)}R_j.
\]
The same advantage weights all sampled actions in an episode, without group-standard-deviation division or batch whitening. Equal-return groups have zero advantage. A variant additionally sets all-success and all-failure groups to zero, even when success bonuses differ; the historical batches do not establish which variant was used in every case. Thus differing success bonuses alone do not establish a learning signal.

\paragraph{Episode weighting.}
Let $m_{it}$ mark valid generated action tokens, excluding prompts and padding, and $T_i=\sum_t m_{it}$. Let $k_{it}$ indicate whether ClipCov retains the policy term, and $c_{it}=\min(s_{it}A_i,\operatorname{clip}(s_{it},1-\epsilon,1+\epsilon)A_i)$. Equal episode weighting gives
\[
 L_{\rm policy}=-\frac1N\sum_{i=1}^N
 \frac{\sum_t m_{it}k_{it}w_{it}c_{it}}{\max(T_i,1)}.
\]
Here $N$ is the number of actual episodes. Each episode averages its valid action-token terms; an episode with no decision contributes zero but remains in $N$. ClipCov removes selected numerator terms without changing $T_i$. Entropy and reference-KL terms are reduced separately. This weights episodes equally rather than pooling all batch tokens.

\paragraph{Relation to Dr. GRPO and DAPO.}
Dr. GRPO removes group-standard-deviation scaling and realized-response-length normalization~\citep{liu2025understandingr1zero}. Our no-standard-deviation choice matches the first convention, while the episode loss retains within-episode averaging. DAPO pools token losses over the total token count~\citep{yu2025dapo}, unlike equal episode weighting. These are individual optimization choices rather than adoption of either complete method.

\subsection{Validation checkpoints}\label{app:ascii-resume-extension}
Checkpoint 335 clears 78/96 normal-start T7 attempts. Its T1--T7 counts are 96, 96, 79, 96, 96, 90, 78 out of 96 each. The eight T7 layouts with twelve samples each also support monitoring and checkpoint selection, so this is validation performance rather than an independent test.

The restart-distribution analysis uses a separate training sample from the plotted continuation: its batch 333 clears 192/192, whereas the curve's batch 333 clears 180/192. The eleven-state concentration results refer only to the former sample. Both use mixed training starts and normal-start validation.

\paragraph{Curriculum map examples.}
Figure~\ref{fig:historical-tier-mazes} shows one validation layout per stage. Training uses 32 catalog layouts and validation eight separate layouts. The illustration depicts normal starts rather than saved restart states. Geometry, objectives and ghost count change together across stages.

\paragraph{Initial and trained policy comparison.}\label{app:historical-endpoints}
Checkpoint 0, before this run's policy optimization, clears 0/96; checkpoint 270 clears 77/96 (80.2\%). Each uses eight held-out T7 layouts with twelve samples per layout, matched initial observations, environment seeds and policy-sampling seeds. Conditions are three extra lives, normal ghosts, 512 environment actions, no restart or action prefix, four-direction masking, one output token, temperature 0.7 and top-$p$ 1.0. A clear collects all ordinary pellets. Later observations diverge as actions differ.

The comparison measures improvement of a training recipe, not an isolated curriculum effect. Incomplete per-sample training-map information limits verification of train/validation separation; repeated training seeds and an independent reporting set would strengthen robustness claims.

\section{Step 2: Visual Grounding and Adaptation}
\subsection{Visual-agent setup}\label{sec:step-two}
Step 2 introduces screenshots and a higher-level action interface. The visual agent selects from a menu of candidate routes, and the harness executes the selected route through primitive moves. Engine-state assistance supplies the harness with game state for candidate generation and risk checks. We first evaluate visual adaptation, then hold the adapted model fixed to compare control mechanisms. Input routing and the engine-state observation condition are detailed in Appendix~\ref{app:environment-interface}.

\subsection{Staging visual grounding and language adaptation}
We adapt the vision encoder and merger while freezing the language component, then freeze vision and train language-only LoRA~\citep{hu2022lora} on visual questions. The first stage targets screenshot grounding; the second adapts the answer interface while retaining the learned visual representation. Training labels describe visible map contents and entities, including occluded cells marked unknown. The image, prompt and output schema are shared across training and evaluation; the reproducible adaptation recipe is given in Appendix~\ref{app:d5a-reconstruction}.

Table~\ref{tab:visual-adaptation-stream} compares historical checkpoints on the same level-1 screenshot stream. The single-size visually adapted model obtains 256/531 exact full-map readings (48.2\%); multi-size visual adaptation obtains 401/531 (75.5\%); adding language-only LoRA gives 427/531 (80.4\%). This progression motivates training visual grounding before adapting the language response. The design aims to reduce reliance on training-layout shortcuts while directing visual learning into the encoder and merger. The scores measure schema-constrained whole-map transcription; the separate raw-generation and interactive reading panels are retained in Appendix~\ref{app:perception-panels}.

Table~\ref{tab:visual-adaptation-stream}A reports the historical stream comparison in the main text. Single-size vision is adapted to the $25\times 21$ size; multi-size vision is evaluated before and after language-only LoRA.

\label{app:method-detail}
\paragraph{Adaptation design.}
The vision stage learns grid and entity transcription while language weights remain fixed. Language-side LoRA then adapts visual question answering with vision fixed. This allocation directs the first stage toward visual features; unseen-layout evaluation is still needed to test reliance on layout shortcuts.

\subsection{Perception SFT and Image Representation}\label{app:d5a-reconstruction}

The reproducible two-stage recipe below uses Qwen3.5-9B and screenshots generated from the game. It describes how to reproduce the adaptation approach; the complete original training datasets for the measured multi-size checkpoints, with and without language LoRA, are unavailable.

Native screenshots use 16-pixel cells. The shared image function applies exactly $2\times$ bicubic enlargement, producing 32-pixel cells, with processor limits of 4,096 and 537,600 pixels. The same preprocessing is used for SFT, evaluation and serving, targeting one visual token per board cell for boards up to $25\times 21$. The prompt's board coordinates remain native grid coordinates after enlargement. The system instruction is ``You answer questions about the visible Pacman screenshot precisely.'' The user prompt specifies board dimensions, coordinate axes and symbols: wall, ghost-house door, ordinary/power pellets, Pac-Man, three ghost states, empty and hidden cells. It requests JSON only. Hidden labels describe pixel occlusion, including sprites and the HUD; they do not reveal objects behind an occluder. The language-model chat template disables thinking, and cross-entropy is computed only on answer and end-of-sequence tokens; image/prompt and padding tokens are masked.

\begin{table}[t]
\centering\small\setlength{\tabcolsep}{3pt}
\caption{Two-stage visual-adaptation recipe. Final planned checkpoints, rather than minimum validation cross-entropy, are selected. }
\label{tab:d5a-recipe}
\begin{tabular}{p{.30\columnwidth}p{.28\columnwidth}p{.28\columnwidth}}
\hline
Setting & Vision stage & Language stage \\
\hline
Trainable part & Encoder and merger & Language LoRA only \\
Updates / batch & 6,000 / 16 & 1,000 / 16 \\
Peak learning rate & $10^{-5}$ & $10^{-4}$ \\
LoRA configuration & Not used & Rank 16, alpha 32 \\
LoRA dropout / bias & Not used & 0.05 / none \\
Optimizer & AdamW & AdamW \\
Adam betas & $(0.9,0.999)$ & $(0.9,0.999)$ \\
Weight decay / clip norm & 0.01 / 1 & 0.01 / 1 \\
Schedule & 10\% warmup, cosine & 10\% warmup, cosine \\
Final rate multiplier & 0.1 & 0.1 \\
\hline
\end{tabular}
\end{table}

The vision implementation uses four-way FSDP2, FP32 master parameters and AdamW states, BF16 computation and FP32 gradient reduction. The language stage accumulates 16 single-example microbatches on one GPU; all vision parameters and non-language adapters are explicitly frozen after adapter construction. Vision and language data use separately seeded sampling schedules, distinct from model initialization. Reconstruction test families include two native levels excluded from training and unseen procedural $25\times 21$ maps; the size panel covers 18 sizes from $11\times 11$ through $25\times 21$. The shipped-map stream panel tests unfiltered map transcription. The resulting scores measure VQA and whole-map transcription.

\subsection{Frozen Perception Prompts and Minimal Outputs}
\label{app:frozen-perception-prompts}

The perception prompt contains one image, the common instruction below and a task suffix. The system instruction is given in Appendix~\ref{app:d5a-reconstruction}.

\begin{quote}\small
The image is a 25-row by 21-column board of 16-pixel cells, starting at image top-left. Grid coordinates are 0-based [row,col], row downward, col rightward. Pixel coordinates [x,y] are absolute image pixels, never grid coordinates. Symbols: \# wall tile, - red ghost-house door, . visible small pellet, o visible power pellet, P Pacman, G normal ghost, V vulnerable blue/white ghost, E eyes-only ghost, \_ empty. A character is written at the cell holding most of its visible sprite; if two characters share a cell, write the one you see on top. ? means hidden: something drawn over the cell (a character, fruit, score text or the bottom score/lives bar) covers it so its contents cannot be seen. A blinking power pellet absent in this frame is empty. Portals look empty. Empty cells are underscores; spaces only separate cells. Return only the requested JSON, without explanation.
\end{quote}

\paragraph{Initial-map suffix.}
\begin{quote}\small
Transcribe this entire visible board. Output exactly \{``rows'':[25 strings]\}, each string exactly 21 symbols separated by single spaces. No indices.
\end{quote}

The quotation marks above denote ordinary JSON double quotes. The decoder generates greedily, with a 2,048-new-token cap, and requires an end-of-sequence token. The input cap is 4,096 tokens. The parser accepts only the key \texttt{rows}; a full-board answer must contain exactly 25 strings, each of length 41 with exactly one standard space between adjacent symbols. Only \texttt{\#-.o\_PGVE?} is accepted. A compact row such as \texttt{\#\#\#} fails even if its intended symbols are recognizable.

\paragraph{Actor suffix.}
\begin{quote}\small
Locate all visible characters, including any partly hidden behind another. Schema: \{``pacman\_rc'':[row,col] or null,``ghosts'':[\{``rc'':[row,col],``state'':``normal|\allowbreak{}vulnerable|\allowbreak{}eyes''\}]\}; ghosts sorted by row then col.
\end{quote}

Minimal illustrative JSON follows; the three-by-three map demonstrates serialization only, using the parser's dimension arguments set to three.
\begin{verbatim}
{"rows":["# # #",
         "# P .",
         "# o #"]}

{"pacman_rc":[1,1],"ghosts":[
 {"rc":[1,2],"state":"normal"}]}
\end{verbatim}

Unspaced maps, out-of-board coordinates and unrecognized ghost-state strings are invalid. Invisible characters remain absent rather than being inserted to meet an expected count. Agent decisions use one advertised option token, not JSON; the code-to-route mapping changes with the menu. The visual policy receives the screenshot, estimated state and candidates, with uncertainty made explicit.

\subsection{Perception Evaluation}\label{app:perception-panels}
\subsubsection{Static screenshot reading}
The frozen visually and language-adapted model reads 24 distinct screenshots from four execution seeds on one $25\times 21$ map. Six approximately equally spaced points per episode are selected on a 16-action sampling grid before inspecting model outputs. They span 192 to two remaining ordinary pellets, vulnerable/eyes sprites, overlap and HUD occlusion. These are new views of one familiar layout.

Visible-pixel labels use native $336\times 400$ RGB, followed by $2\times$ bicubic enlargement for the model. Private targets are excluded from prompts. Full maps are generated as strict single-space JSON, without constrained grid decoding or cell repair. Each image receives one greedy output and three temperature 0.7 samples; one initial image receives ten additional greedy repetitions. All 106 outputs end normally.

Table~\ref{tab:perception-units} collects these static and interactive measurement units.

The three sampled errors are two boundary walls transcribed as empty and a visible ghost transcribed as the ghost-house door: 37,797/37,800 cells are correct. Per-cell majority over the three sampled outputs is exact on all 24 images, but uses the same bank and extra calls; it is not an independent reliability test. The ten initial greedy outputs have one distinct text and token sequence. Repeated determinism is not additional layout coverage.

\subsubsection{Closed-loop reading}
Three execution seeds each produce a zero-extra-life game with a 64-action window. The reader requests a full map initially and every eight actions, and actor positions between. Executed lengths are 64, 30 and 64 because one game ends early. Of 158 correlated observations, 18/20 maps, 131/138 visible-ghost multisets, 138/138 player positions and 137/138 ghost counts are exact. Both map errors assert empty floor at an occluded cell; unknown cells should remain uncertain. Surviving a window is not full-game completion.

\subsubsection{Additional visual-adaptation comparisons}
Schema-constrained transcription for multi-size vision without versus with language LoRA is 401/531 versus 427/531 on the screenshot stream, 50/64 versus 51/64 on an unseen family, and 182/216 versus 189/216 on unseen sizes. These checkpoint measurements are separate from the raw-generation image bank and from newly reproduced training.

\section{Step 2: Candidate-strategy Harness and Control}
Across model adaptation and harness improvements, the Step 2 clear rate rises from 26.6\% to 96.9\% on one fixed map with three extra lives: Qwen with target replanning and the basic gate clears 17/64, while the adapted model with route binding and the improved gate clears 62/64 (Table~\ref{tab:factorial-complete}). The two endpoints change both model and control configuration; the fixed adapted-model comparison below starts at 40/64.

\subsection{Harness Details}
\paragraph{Options.} At each decision the harness offers up to \pend{4} \textsc{Collect} options, \pend{4} \textsc{Avoid} options, and \pend{2} \textsc{Eliminate} options, each selected through an advertised single-letter code whose target changes with the state. After the model picks a code, \edward{} converts the target into U/D/L/R moves, re-validates the option at every step, and calls the model again when the target is reached, the option becomes invalid, or a step limit is hit.
\paragraph{Information.} In the engine-state condition the harness reads Pac-Man and ghost positions and states from the engine at every step, so that condition is not purely visual. The factorial route gates and shared fallback rule are specified in Appendix~\ref{app:factorial-protocol}; the public-memory execution sketch is in Appendix~\ref{app:public-feedback-contract}.

\paragraph{Risk checking and route commitment.}
The VLM selects state-dependent Collect, Avoid and Eliminate routes using single-letter codes. Candidate checks estimate whether a ghost can reach the route first over a bounded horizon. Route binding executes the selected sequence until completion or interruption; target replanning recomputes a route toward the selected target with limited commitment. Both reuse model choices, following temporally extended options~\citep{sutton1999options}. When no acceptable route remains, refusal stops execution, while risk-ranked fallback continues with the least risky available action. Each comparison specifies its fallback rule.

\subsection{Fixed-weight harness comparison}\label{sec:harness-result}
The harness offers state-dependent \textsc{Collect}, \textsc{Avoid} and \textsc{Eliminate} routes, represented by advertised single-letter codes. In the control panel, the adapted model's weights stay fixed while a $2\times 2$ comparison changes execution commitment and route checks. Each arm uses 64 paired execution seeds on one canonical $25\times 21$ map, four normal ghosts, no fruit, temperature 0.7 and a 512-native-step limit. The 0- and 3-extra-life settings are reported separately.

The target-replanning control keeps the selected target but recomputes its route after each move, asking the model again after at most 8/3/6 moves for Collect/Avoid/Eliminate. Bound execution instead continues the selected route until completion, invalidation or safety interruption. The improved gate adds a two-move motion-envelope check and a near-target escape guard. All four arms share the candidate-generation scaffold and fallback; Appendix~\ref{app:factorial-protocol} specifies these mechanisms and the full two-model panel.

Table~\ref{tab:factorial}B reports the four fixed adapted-model arms in the main text. Model weights, proposal scaffold and fallback are fixed. The target-replanning control shares the factorial multi-route scaffold and differs from the original Route-option planner.

With three extra lives, jointly changing historical acceptance with target replanning to improved acceptance with bound execution raises completion from 40/64 (62.5\%) to 62/64 (96.9\%). The corresponding 0-extra-life counts rise from 7/64 to 27/64. The separate contrasts locate the gain more precisely: binding under the basic gate changes 7/64 to 9/64 and 40/64 to 39/64; under the improved gate it changes 18/64 to 27/64 and 59/64 to 62/64. Binding's benefit therefore depends on the route gate and life budget.

This experiment shows how much realized capability can change through control design with learned weights fixed. The model's choices interact with which routes the harness admits and how long it executes them. In the complete panel, binding under the improved gate changes unadapted Qwen from 11/64 to 9/64 with zero extra lives and leaves 37/64 unchanged with three. Model-specific completion, deaths and matched-success call costs appear together in Appendix~\ref{app:factorial-protocol}.

\subsection{Harness Factorial: Design and Model Dependence}\label{app:factorial-protocol}
The factorial includes 1,024 episodes: two models, two life budgets, four arms and 64 paired execution seeds per cell. All use one $25\times 21$ map, four normal ghosts, no fruit, temperature 0.7 and 512 native steps. Each arm starts from matched random seeds; subsequent trajectories can diverge as actions differ.

\paragraph{What changes.} Target replanning maintains the selected target, recomputes its route after every move, and caps commitment at 8/3/6 moves for Collect/Avoid/Eliminate (one for fallback). Route binding binds the selected route and continues its recorded sequence until completion, invalidation or safety interruption. The execution choice therefore changes both execution commitment and the timing of subsequent model calls. The basic risk check applies the earlier per-strategy route acceptance and continuation gates; The improved risk check applies a two-move motion-envelope check and a near-target escape guard. For Collect targets within two moves, the latter searches up to four additional moves for an escape to a junction and checks the resulting path. The checks cover these bounded motion and escape horizons.

\paragraph{What stays fixed.} All arms use the same geometric multi-route Collect proposals, the same native Avoid/Eliminate proposals with their existing pre-screen, candidate ordering, visit penalty, route summaries, public-kinematics risk-ranked fallback and escape hints. Risk checking compares route eligibility/continuation within this shared, pre-screened scaffold. Target replanning with basic risk checks differs from the original route-option planner. Target-replanning prompts truthfully describe target replanning; bound-route prompts describe bound execution. The original planner with risk-ranked fallback is a separate reference condition.

\begin{table}[ht]
\centering\small\setlength{\tabcolsep}{4pt}
\caption{Complete fixed-map factorial. Each entry gives clears/64; death events appear in parentheses. Adapted denotes multi-size visual adaptation plus language LoRA. Each row names its execution and risk-check conditions.}
\label{tab:factorial-complete}
\begin{tabular}{llrr}
\toprule Model & Execution / check & 0 extra lives & 3 extra lives \\
\midrule
Qwen & \shortstack[l]{Target replan\\basic} & 2/64 (44) & 17/64 (114) \\
Qwen & \shortstack[l]{Target replan\\improved} & 11/64 (35) & 37/64 (49) \\
Qwen & \shortstack[l]{Bound route\\basic} & 3/64 (50) & 27/64 (113) \\
Qwen & \shortstack[l]{Bound route\\improved} & 9/64 (34) & 37/64 (58) \\
Adapted & \shortstack[l]{Target replan\\basic} & 7/64 (46) & 40/64 (92) \\
Adapted & \shortstack[l]{Target replan\\improved} & 18/64 (35) & 59/64 (55) \\
Adapted & \shortstack[l]{Bound route\\basic} & 9/64 (52) & 39/64 (108) \\
Adapted & \shortstack[l]{Bound route\\improved} & 27/64 (32) & 62/64 (57) \\
\bottomrule
\end{tabular}
\end{table}

Under improved risk checks, binding changes unadapted Qwen clears from 11 to 9 (0 extra lives) and 37 to 37 (3 extra lives), compared with 18 to 27 and 59 to 62 for the adapted model. On paired episodes that both arms clear, mean calls decrease by 43.67/46.40 for Qwen and 47.88/26.91 for the adapted model (0/3 extra lives). The matched-success subsets have $n=3$ and $20$ for Qwen, and $n=8$ and $57$ for the adapted model, for 0 and 3 extra lives respectively. 3-extra-life deaths nevertheless rise from 49 to 58 for Qwen and 55 to 57 for the adapted model. The call-cost comparison applies to these jointly successful episodes.
\subsection{Independent Checkpoint Evaluation and Latency}
\label{app:independent-profile}
A separate fixed-map evaluation compares base Qwen3.5-9B with three RL checkpoints after 19, 48 and 49 updates. It evaluates five conditions, including a separate direct-action base condition, with twelve games each: four execution seeds, three generation seeds per execution seed, one fixed maze, normal ghosts, 0-extra-life termination and a 512-environment-step cap. The model receives real images, generates one legal token, uses temperature 0.7/top-p 1 and disables thinking. Route-option harness conditions retain the saved screenshot-plus-structured-live-state/history interface and refusal fallback. Direct versus Route-option harness changes both information and control.

The update 19 checkpoint is selected by mean training reward, not held-out completion. Its preceding training batch clears 22/48; those preceding updates 48 and 49 clear 12/48 and 10/48. These training samples are separate from the twelve-game evaluation.

\begin{table}[ht]
\centering\small\setlength{\tabcolsep}{4pt}
\caption{Independent profile outcomes, twelve attempts per condition. Refusals are harness safety stops, not service errors; they count as non-clears. All completed evaluations are included. Here harness denotes the Route-option harness; base identifies the base model and update labels identify the saved RL checkpoints.}
\label{tab:independent-profile}\label{tab:headline}
\begin{tabular}{lrrrr}
\toprule Condition & Clears & Deaths & Refusals & Limit \\
\midrule
Direct base & 0 & 12 & 0 & 0 \\
Base + harness & 4 & 0 & 7 & 1 \\
RL update 19 + harness & 3 & 1 & 8 & 0 \\
RL update 48 + harness & 3 & 0 & 9 & 0 \\
RL update 49 + harness & 2 & 0 & 8 & 2 \\
\bottomrule
\end{tabular}
\end{table}

These checkpoints come from separate policy training, rather than output-head adaptation of the visual-language model.

\paragraph{Serving and timing.} Each model uses one H800 80 GB GPU, TP=1, BF16, vLLM 0.22.1/MRv1 eager execution, 1,024-token context and one concurrent request per service; prefix and multimodal processor caching are disabled. Eight warmup requests precede each seed block and are excluded.

The timing protocol is stated explicitly for comparison with timing-sensitive game evaluations~\citep{tang2026realtimebench}. During these evaluations, the game clock waits for model inference and advances when actions execute.

Mean one-token request time is 115.04\,ms for direct base and 114.60\,ms for the base model with the Route-option harness, including image processing, prefill, generation and server-local HTTP. The observation-return-to-action-start pause, including harness preparation and the model request, averages 169.11\,ms for direct base control and 271.12\,ms for the base model with the route-option harness. The base model with the Route-option harness executes 2,433 actions over 1,330 decisions (1.829 actions/decision), spreading its decision cost across multiple actions and yielding the approximately 148\,ms amortized decision pause reported in Appendix~\ref{app:finding-detail}. These mean timings quantify sub-second request and decision costs, with lower amortized decision pause in the options condition. The pause excludes game advancement/rendering inside the environment call, option reuse, model startup/warmup and the Windows-to-server network round trip. Full step wall time was not measured. A separate 64-request identical-image/prompt bank is replayed on all four model identities to profile serving differences.

\subsection{External-model baselines}\label{app:external-baselines}
GPT-6 Luna Decisions API runs used 50 seeds, one layout, normal ghosts, no fruit, 0 extra lives and 512 game steps. Native prompts included structured state. Image Route-option harness used refusal fallback and had 3 API refusals; text-based maze Route-option harness used risk-ranked fallback and had 1. Failures count as non-clears, without retries. The modality comparisons therefore also differ in fallback.

A separate exploratory Codex CLI game uses GPT-6.1 Sol with text-based maze input and an experimental route-option harness. It clears 1/1, scores 11,920 in 386 steps and 129 model calls, with no deaths and no ordinary or power pellets left (Table~\ref{tab:codex-early}). This is one preliminary attempt.

\begin{table}[t]
\centering
\small\setlength{\tabcolsep}{4pt}
\begin{tabular}{llrrr}
\toprule
Input & Agent & Clears & Win rate & Mean score \\
\midrule
Image & Primitive & 0/50 & 0\% & 490.2 \\
Image & \shortstack[l]{Route-option\\harness} & 10/50 & 20\% & 5104.5 \\
\shortstack[l]{Text-based\\maze} & Primitive & 0/50 & 0\% & 242.6 \\
\shortstack[l]{Text-based\\maze} & \shortstack[l]{Route-option\\harness} & 14/50 & 28\% & 5571.0 \\
\bottomrule
\end{tabular}
\caption{External baseline: GPT-6 Luna via Decisions API; each input includes structured state. Fixed layout, 50 seeds, 0 extra lives, no fruit, 512 game steps. Win rates include API failures as non-clears; mean scores use completed episodes (50, 47, 50, 49). Image Route-option harness uses refusal fallback; text-based maze Route-option harness uses risk-ranked fallback.}
\label{tab:external-baseline}
\end{table}

\begin{table}[t]
\centering
\small
\begin{tabular}{lrrr}
\toprule
Model / interface & Clears & Score & Game steps \\
\midrule
GPT-6.1 Sol / Codex CLI & 1/1 & 11,920 & 386 \\
\bottomrule
\end{tabular}
\caption{Preliminary single-game result: text-based maze input, experimental Route-option harness, low reasoning effort. The game cleared all normal pellets with zero deaths in 129 model calls. The 1/1 outcome describes this single attempt.}
\label{tab:codex-early}
\end{table}

\label{app:external-baseline-interface}

\section{Step 3: Observation-driven Memory and Preliminary Diagnostics}
\subsection{Perception accuracy depends on the measurement setting}\label{sec:perception-result}
A frozen-model evaluation reuses the original training prompt and visible-pixel labels on 24 distinct screenshots: six temporal samples from each of four episodes on one canonical map. Greedy raw generation transcribes 24/24 exactly, while three samples per image at temperature 0.7 are exact on 69/72; all outputs have valid strict JSON/spacing. Repeating one initial image ten times yields identical correct text and tokens, a stability check rather than ten independent images.

During three separate 0-extra-life 64-action diagnostic windows, the hybrid reader makes 20 full-map queries (18 exact) and 138 actor queries (131 exact visible actor sets). Pac-Man is exact on all 138 actor queries; the remaining errors concern ghosts. Only 158 actions are executed because one episode dies at action 30. The 158 observations are correlated within the three windows. The map errors each replace an occluded unknown cell with asserted empty floor, showing why valid formatting is insufficient. Static and closed-loop panels use different images, decoding and query types. Prompt/schema alignment is one check for interface mismatch.

\begin{table}[ht]
\centering\small\setlength{\tabcolsep}{4pt}
\caption{Separate perception tasks and denominators. The first three rows reuse the same 24-image bank; the last two come from three short game windows. Rows measure reading accuracy.}
\label{tab:perception-units}
\begin{tabular}{p{.52\columnwidth}rr}
\toprule Query panel & Valid & Exact \\
\midrule Raw greedy, 24 images & 24/24 & 24/24 \\
Raw sampled, three per image & 72/72 & 69/72 \\
Initial-image greedy repeats & 10/10 & 10/10 \\
\midrule Periodic full-map queries & 20/20 & 18/20 \\
Actor queries & 138/138 & 131/138 \\
\bottomrule
\end{tabular}
\end{table}

Table~\ref{tab:perception-units} separates response validity from exact visible-state content. A valid answer obeys the output format; exactness additionally requires agreement with the visible-pixel labels, including unknown cells. The same image bank supplies the first three rows, so 72 samples and ten repeated decodes do not provide 82 additional layouts. Correct Pac-Man locations on all 138 actor queries also do not certify complete actor readings: the exact visible-actor sets number 131.

In the static sampled outputs, 37,797 of 37,800 cells are correct, yet three complete-map answers are wrong. The observed errors include boundary walls read as empty and a ghost read as the house door. An almost perfect cell average can therefore conceal an incorrect board-level interpretation. Conversely, repeated greedy agreement measures decoding consistency on the same image, not success over changing observations.

The relevant distinction is between recognizing the visible scene and producing a response that the decision interface can use. Static-map transcription, actor readings and output formatting exercise different parts of this interface. These observations motivate separate checks of visible-pixel semantics and the actual prompt and parser format before interpreting a game failure as a policy-learning problem.

This is why we evaluate perception separately from the text-based maze learning curve. The screenshot records locate concrete reading problems; the textual policy records ask what happens when image transcription is removed.

\subsection{A memory-overwrite case}
\paragraph{From a reading error to persistent state.}
In an early comparison, engine assistance clears 28/64 games with zero extra lives and 57/64 with three; the screenshot-feedback prototype clears 0/64 in both settings, with 14 and 34 harness aborts. One observed failure places Pac-Man at $[17,17]$ instead of $[17,16]$. Treating the occupied cell as passable then overwrites a correct wall with floor (Figure~\ref{fig:memory-audit}). This identifies an error-propagation mechanism, but the comparison also differs in other system components.

The case is illustrated in Figure~\ref{fig:memory-audit}.

\paragraph{Reading and state-update error.}\label{app:historical-memory-trace}
The actor query gives the correct player cell but returns five ghost detections and is rejected. The fallback full-map query places the player one cell to the right. That location triggers the terrain overwrite, which persists when the next player reading becomes correct. The failure is therefore a combination of fallback perception and an update rule that lets actor estimates alter static terrain.

The same scene contains a ghost overlapping a wall near $[20,10]$. The cause of this engine anomaly is unresolved, so the example does not isolate image-reading failure or quantify its contribution to game outcomes. This early prototype receives changed-cell, pickup, score, life and termination events, but no relative-displacement feedback; the later memory system below has a richer interface.

With static terrain separated from actor estimates, two fixed-action windows retain correct player positions and pellet sets for 64 and 57 steps in the zero- and three-extra-life conditions. The latter stops after a five-ghost response. These windows support the revised update rule locally; freezing terrain alone would preserve any initial map error, and full-game comparisons remain necessary.

A separate 64-seed paired full-game comparison on one fixed map uses the same visually and language-adapted policy. Engine assistance clears 26/64 and 60/64 with zero and three extra lives; public feedback clears 19/64 and 55/64. Public-feedback deaths are 42 and 93 versus 32 and 58, with 16,420 and 25,810 additional perception calls. Player, pellet and clock estimates are correct, but ghost omissions and location errors remain. The public interface supplies relative movement, events and known rules. These are whole-system, fixed-map comparisons with unequal compute, separate from the early prototype above.

The revised public-feedback variant separates updates by type. Its \emph{static map} comes from the initial visual read and retains unknown cells; \emph{pellets} change through public pickup events; and \emph{dynamic actors} are re-read from screenshots. Actor--map conflicts do not overwrite terrain. Visually observed spawn points and public teleport/respawn rules handle discontinuous moves. These are revised update rules, not rules retroactively shared by all historical harnesses. The next test is a matched full-episode comparison of the original and revised rules.

\subsection{Visible Inputs and Public-feedback Memory}
\label{app:public-feedback-contract}

The later public-feedback system transcribes the map once, then reads visible actors after actions. Static terrain, ordinary pellets, power pellets and actor estimates are maintained separately. Coordinates are zero-based rows and columns, increasing down and right. Hidden cells remain unknown and engine truth never fills them. This variant differs from the earlier prototype by including relative movement feedback.

Public interaction feedback supplies elapsed logic ticks, relative player motion, localized pickups, deaths, respawns, termination and eaten-ghost events. It excludes hidden maps, absolute ghost coordinates, identities, paths and private timers. Pickups update pellet memory before death or respawn changes player position; uncertain contacts remain uncertain. Death resets vulnerability memory, respawn uses the visually observed spawn, and motion propagates only onto known floor. Public power events and the 360-tick rule determine a vulnerability interval.

The actor query asks for all visible characters, including partial occlusions, with a Pac-Man coordinate or null and a list of ghost coordinates and states (normal, vulnerable or eyes), sorted by row and column. The parser preserves extra detections and multiplicity. Temporal ghost association minimizes Manhattan displacement over possible assignments; ambiguous ties remain uncertain. Its per-observation displacement bounds are two cells for normal/vulnerable ghosts and four for eyes. These are visual estimates, not persistent engine identities or guaranteed next-step headings. Existing options continue only while their public validity checks hold; a new Agent decision is required when an option is absent or invalidated.

\paragraph{Evaluation and perception failures.}
Private game state scores field accuracy but is excluded from policy inputs and learning labels. Invalid observations stop the affected attempt as a perception failure with undefined game reward, rather than being counted as an ordinary death. Correct player or pellet memory does not imply correct ghost estimates.

\paragraph{Public-memory control loop.}
The system parses the visible initial map, reads actors, updates public memory and either continues a valid option or requests a new route choice. After primitive execution it incorporates public feedback and repeats until termination or the action cap. Ambiguous fields remain unknown throughout.

\subsection{Public-memory Fields, Motion Inputs, and Diagnostic Scope}
\label{app:public-memory-motion}
\label{app:public-memory-followup}

The following experiments use frozen visually and language-adapted weights with screenshot-derived motion estimates in the later public-feedback system. They measure full-game outcomes, field accuracy and local gate sensitivity separately.

\paragraph{Feedback detail matters.}
Localizing a pickup relative to the previous player position supplies more information than a Boolean pellet-eaten signal. Processing it before death/respawn avoids assigning the pellet to a new spawn location. Uncertain contacts remain unknown; exact ghost timers are not supplied.

Holding actions fixed across three trajectories of 64, 173 and 168 steps, the earlier pellet memory differs from truth in 0, 80 and 159 frames. Revised feedback and updates eliminate pellet-set errors across all 405 actions, including three pickup-plus-death transitions. All 396 applicable nonterminal, non-death timer intervals contain the true value. This tests the feedback and update rule jointly, not new closed-loop decisions.

\paragraph{Player, pellet and timer accuracy.}
Four full games cross two execution seeds with zero and three extra lives, capped at 512 actions. All end by death or life exhaustion, with no clears, timeouts or technical faults (Table~\ref{tab:public-memory-fullgames}). Player positions and pellet sets are exact across 431 pre-action observations; all vulnerability intervals contain the true timer. The games use 383 model choices and 431 perception calls. Ghost accuracy is measured separately.

\begin{table}[t]
\centering\small\setlength{\tabcolsep}{3pt}
\begin{tabular}{lrrrr}
\toprule
Extra lives & Steps & Deaths & Pellets left & Agent calls \\
\midrule
0 & 83 & 1 & 160 & 73 \\
0 & 30 & 1 & 192 & 30 \\
3 & 180 & 4 & 130 & 157 \\
3 & 138 & 4 & 144 & 123 \\
\bottomrule
\end{tabular}
\caption{Four terminal public-memory games on one fixed map; none clears. Remaining ordinary pellets are private outcome measurements. The same two seeds are listed in matching order under each life budget. These four games are not independent maps or a stable win-rate estimate.}
\label{tab:public-memory-fullgames}
\end{table}

At least one normal/vulnerable ghost has unknown estimated heading in 291/431 frames. Position-only association matches 1,627 ghost observations: 1,599 at zero Manhattan error, 26 at one cell and two at two cells; another 97 true-ghost observations are unmatched and there are no extras. These are ghost observations across correlated frames, not 1,627 independent entities. Only 48 actions continue an already bound route; the others require new Agent choices. The frozen short-horizon gate returns false when an active ghost has unsupported heading input, even before excluding a distant irrelevant ghost. A false result can therefore mean ``insufficient motion information'' rather than a predicted collision.

\paragraph{Selected same-state input replacements.}
The retained private diagnostic selects the first three snapshots per game with four distinct ghost cells, exact position/state association and at least one unknown active heading. In all four games these are steps 0, 4 and 5; repeated early states and this selection criterion limit the sample. At each of these twelve snapshots the visible map, player, food and frozen safety function remain fixed. With observed inputs every open direction fails the one-step gate. Replacing only ghost headings with private truth makes at least one direction pass in 12/12; replacing only subcell ghost pixels does so in 0/12; replacing both does so in 12/12. This identifies an input precondition in these selected local checks. The measured endpoint is one-step gate acceptance. Oracle fields never enter the deployed policy.

\paragraph{A separate bounded heading-reading probe.}
Twelve pre-fixed dynamic states come from one trajectory, including normal, vulnerable and eyes states. Each receives one greedy current-image query and one previous/current-image query: 24 model calls, all valid JSON with natural EOS and no truncation. Headings are an added task, not the position/state target of the original adaptation. Scoring associates by position alone; only assignment pairs unique across minimum-distance solutions and within one Manhattan cell are eligible. True heading cannot break a positional tie. Table~\ref{tab:heading-probe} retains all 48 true-ghost observations per method, including unmatched/ambiguous cases. Temporal differencing uses existing VLM position readings without extra model calls.

\begin{table}[t]
\centering\small\setlength{\tabcolsep}{3pt}
\begin{tabular}{p{.30\columnwidth}rrrr}
\toprule
Method & Eligible & \shortstack{Correct/\\known} & \shortstack{No\\heading} & \shortstack{Un-\\resolved} \\
\midrule
Current image & 40 & 1/8 & 32 & 8 \\
Two images & 37 & 2/21 & 16 & 11 \\
Position difference & 40 & 23/27 & 13 & 8 \\
\bottomrule
\end{tabular}
\caption{Heading diagnostic on twelve states from one trajectory. ``Known'' means an explicit direction on an eligible pair; ``unknown'' is an eligible pair without one; unresolved is positional/identity ambiguity or failed matching. Each row has 48 true-ghost observations. Conditional accuracy is not complete-state accuracy.}
\label{tab:heading-probe}
\end{table}

The two-image model emits U for all 21 explicit eligible predictions. Temporal direction is more accurate on this probe, but estimates past-interval motion, not guaranteed future heading at a corner. Neither temporal association nor extra images therefore supplies engine-equivalent motion truth.

\paragraph{Matched short windows.}
Using one shared execution seed, privileged-state and revised public-memory control are compared for 64 actions in each life setting, with environment, model and sampling matched. Public player/pellet fields and timer intervals are correct across 128 observations. Both arms survive the windows, with zero deaths under zero extra lives and one under three. Privileged versus public remaining pellets are 140 versus 156 and 152 versus 165; model calls are 37 versus 45 and 40 versus 60. These windows use different seeds and environment settings from the earlier 62/64 panel and do not measure full-game completion.

\paragraph{Gate sensitivity across all observed states.}
For all 431 pre-action states from the four games, we compare estimated motion, oracle headings, oracle subcell pixels and both substitutions. Replacements apply only to uniquely matched active ghosts; ambiguous or unmatched detections remain unchanged. The strict subset requires four distinct estimated and true ghosts with unique exact cell/state matches. Private fields are diagnostic inputs only.

\begin{table}[t]
\centering\small\setlength{\tabcolsep}{3pt}
\begin{tabular}{lrrrr}
\toprule
States & Baseline & Direction & Pixel & Both \\
\midrule
All recorded (431) & 292 & 92 & 292 & 93 \\
Strict matched (321) & 190 & 1 & 190 & 2 \\
\bottomrule
\end{tabular}
\caption{Number of states for which every tested action receives a false gate output. Denominators are recorded states, not games or ghost observations. The strict subset requires exactly four estimated and four native ghosts, distinct cells in each list, and a unique exact one-to-one cell/state match; it is not selected by heading availability.}
\label{tab:all-state-gate}\label{tab:main-motion-audit}
\end{table}

Across all 431 states, direction replacement changes 559 ghost headings; pixel replacement is applied to 1,453 matched active ghost observations and changes 214 pixel fields. These ghost-observation counts differ from the 291 states containing an unknown active heading. Direction-only changes 200 baseline all-false outputs to at least one true output; both replacements change 199. Pixel-only leaves the frame-level classes unchanged. The combined arm is not strictly better than direction alone. The endpoint is local gate sensitivity to missing motion information across correlated states from four games; subsequent trajectories are not evaluated.

\paragraph{Paired motion-prompt pilot.}
Adding public motion information is compared on four fixed-map seeds per life setting (eight pairs, sixteen games). Under zero extra lives both arms clear 0/4; deaths decrease from four to two because two augmented games instead time out. Under three extra lives clears change from 4/4 to 3/4 and deaths from four to eight. Among the three jointly successful pairs, added motion saves nine model calls on average, but total calls over all four games rise from 683 to 691. No technical faults occur.

\paragraph{One-step alternatives at recorded deaths.}
A separate CPU diagnostic replayed each legal one-step alternative at 24 recorded death snapshots: eight had at least one alternative with no immediate death, whereas all open alternatives died in sixteen. The observed-input gate assessed the recorded choice as safe in ten snapshots and the private-truth gate in none. These are selected local checks, distinct from the 431-state input-sensitivity panel and its 321-state strict subset. A one-step survivor is neither a complete counterfactual trajectory nor a measured full-game rescue, and the 8/24 count must not be reported as an episode-level rescue rate.

\subsection{Public-feedback Action-head RL}\label{app:head-rl-protocol}

This experiment freezes the base model, perception and public-feedback harness, and learns only a rank-8, alpha-8 residual at the output head: $z=z_0+(\alpha/r)U(Dx)$. Its 2,019,328 FP32 parameters are initialized with a random down projection and zero up projection. The adapter is disabled for perception calls. This head adaptation is distinct from language LoRA and from the full text-based maze policy training in Section~\ref{sec:system}.

The additional-update protocol uses training batches containing two execution seeds with four complete native episodes per seed (eight rollouts, 0 extra lives, maximum 512 native steps). The 128 training seeds are shuffled deterministically each epoch. The experiment uses normal starts on one fixed $25\times 21$ layout, four normal ghosts and no fruit; it does not use failure-state restarts. Temperature is 0.7 on the legal action support. There is no replacement/backfill of faulty episodes. Four returns from one seed form an advantage group, standardized using the population standard deviation and $10^{-8}$ stabilizer. Constant-return groups receive zero advantage and are skipped.

For a valid episode, the run-specific reward is zero without a clear, and on a clear is
\[
 R=1+0.1(1-d/3)+0.05(1-s/512),
\]
where $d$ is the death count and $s$ the native step count. Faults have no game reward. Other experiments use their own binary terminal reward. At most 32 actual Agent decisions per episode are sampled uniformly without replacement for the update. Episodes receive equal weight; each retained decision receives weight $1/(N L_i)$ for $N$ retained episodes and $L_i$ selected decisions in episode $i$.

The loss uses a PPO-style clipped surrogate (clip 0.2) and categorical KL penalty 0.01 to the frozen base, on the same legal support at temperature 0.7. AdamW uses learning rate $5\times 10^{-4}$, betas $(0.9,0.999)$, epsilon $10^{-8}$, zero weight decay and gradient-norm clipping 1. Behavior-probability consistency tolerance is $2\times 10^{-4}$. Updates require finite, nonzero gradients.

The comparison uses heads after three and twenty total optimizer updates. Eight monitoring seeds, separate from training and paired-comparison seeds, are evaluated under both life settings and are not used to select these reported checkpoints.

\paragraph{Paired comparison.}
Both heads use the same 64 execution seeds under zero and three extra lives on the same fixed map, with a 512-action cap. Each pair shares its initial state and sampling-seed stream. Subsequent trajectories can differ. Table~\ref{tab:head3-head20-protocol-results} reports all outcomes; these seeds vary execution rather than layout.

\begin{table}[t]
\centering\small\setlength{\tabcolsep}{3pt}
\caption{Fixed-map public-feedback action-head comparison. ``Reads'' counts perception calls; deaths count events, so 3-extra-life totals can exceed 64. Each cell contains 64 complete episodes with zero faults.}
\label{tab:head3-head20-protocol-results}\label{tab:head-compact}
\begin{tabular}{llrrrr}
\hline
Extra lives & Updates & Clears & Deaths & Agent & Reads \\
\hline
0 & 3 & 18/64 & 43 & 6,698 & 16,139 \\
0 & 20 & 9/64 & 49 & 6,309 & 15,564 \\
3 & 3 & 55/64 & 101 & 10,941 & 25,628 \\
3 & 20 & 50/64 & 109 & 10,796 & 25,993 \\
\hline
\end{tabular}
\end{table}

On 0-extra-life pairs, five outcomes improve and fourteen regress; on 3-extra-life pairs, eight improve and thirteen regress. Exact two-sided paired-discordance tests give $p=0.0636$ and $p=0.3833$, respectively. This panel shows no observed benefit from the additional updates. Aggregate call reductions accompany more deaths. Among pairs that both clear, the twenty-update head uses 46.75 more steps and 5.5 more Agent calls on average in the four 0-extra-life pairs; in the 42 3-extra-life pairs it uses 6.45 more steps and 3.86 fewer Agent calls. These costs describe the small, jointly successful subsets.

The paired clearance changes are $-14.1$ and $-7.8$ percentage points, with conservative 95\% bounds $[-33.5,+7.3]$ and $[-29.1,+14.4]$. These conditional fixed-map bounds and their assumptions appear in Appendix~\ref{app:paired-uncertainty}.

\begin{figure*}[t]
\centering
\includegraphics[width=0.96\textwidth]{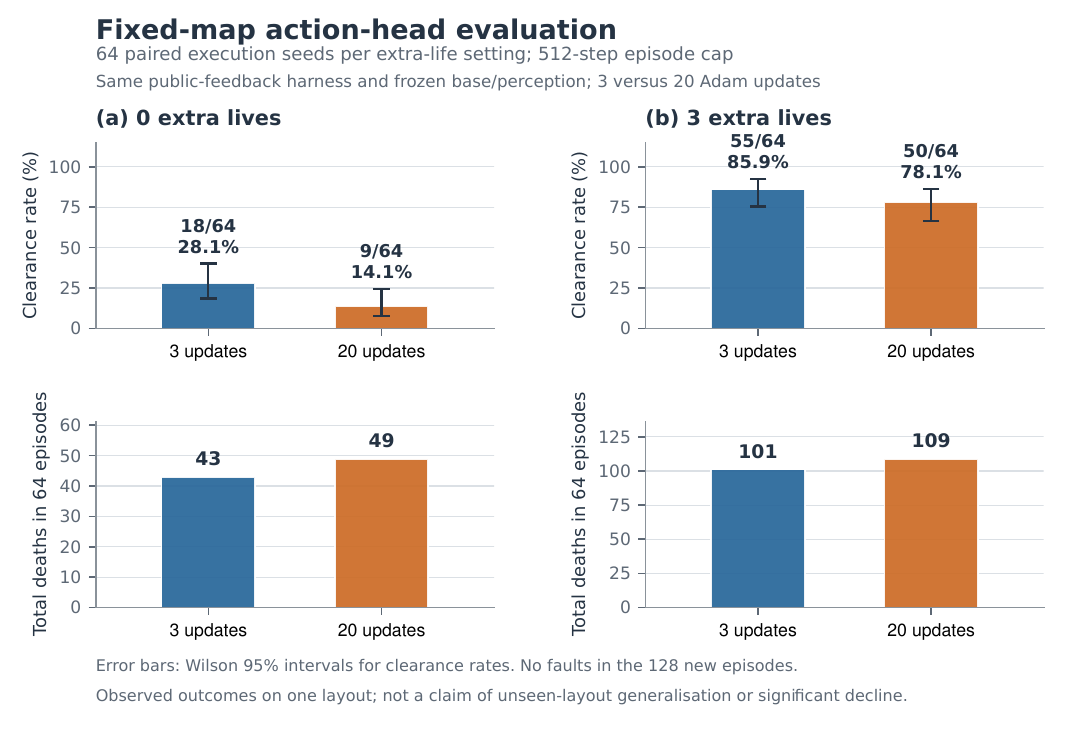}
\caption{Fixed-map public-feedback comparison: three versus twenty total Adam updates of the action head, with frozen base model and perception. Each extra-life setting uses 64 paired execution seeds and a 512-step cap. Upper panels: clear rates with Wilson 95\% intervals. Lower panels: total deaths. The fixed-map differences are not statistically significant.}
\label{fig:head-training}
\end{figure*}

\subsection{A Separate Visual-policy Training Diagnosis}
\label{app:sparse-visual-signal}

This separate public-feedback language-LoRA study uses three extra lives. It analyzes policy checkpoints 4--15 and complete training batches 32--63, separately from residual action-head training and the textual curriculum. Technical faults have undefined rewards and invalidate their original twelve-episode group, without replacement by successful retries.

\paragraph{Policy drift on matched inputs.}
Of fifteen matched initial decision inputs, twelve offer a first avoidance candidate. On those twelve inputs, the mean probability assigned to that individual candidate rises from 15.22\% at checkpoint 4 to 80.10\% at checkpoint 15. Across all fifteen inputs, entropy decreases from 1.277 to 0.795 nats, maximum-option probability increases from 50.80\% to 72.84\%, and mean distribution total variation is 61.33\%. These measure policy change at fixed inputs rather than full-game improvement.

\paragraph{Outcomes at matched pellet density.}
Both periods use 60\% pellet density and the same 36-map collection, but game seeds and trajectories are not paired. Table~\ref{tab:visual-training-eras} summarizes training attempts across multiple checkpoints. Longer games and fewer deaths need not improve completion when timeouts and remaining pellets increase.

\begin{table}[t]
\centering\small
\begin{tabular}{lrr}
\toprule
Recorded statistic & \shortstack{Checkpoints\\4--13} & \shortstack{Checkpoints\\14--15} \\
\midrule
Clears / planned attempts & 57/2,304 & 1/2,964 \\
Technical faults (separate) & 155 & 168 \\
Time limits / valid games & 76/2,149 & 352/2,796 \\
Mean valid-game steps & 248.5 & 301.5 \\
Mean ordinary pellets left & 43.8 & 46.4 \\
\bottomrule
\end{tabular}
\caption{Retrospective visual-policy training outcomes at 60\% pellet density. Faults stay in the planned-attempt denominator for the clear fraction and are separately excluded from valid-game descriptive statistics. These are not independent test episodes or a comparison with the action-head panel.}
\label{tab:visual-training-eras}
\end{table}

\paragraph{Sparse reward and group variation.}
Batches 32--63 contain 6,144 planned attempts and 423 technical faults. Excluding groups with a fault leaves 440 of 512 groups. None has informative within-group advantage under the original success objective. A remaining-pellet proxy varies within all 440 groups, indicating a possible source of denser supervision. Testing such an objective requires a separate matched comparison, without tuning on held-out outcomes.

\paragraph{Probability consistency.}
Across eleven updates, behavior and training distributions differ by mean total variation 3.6--4.3\%. Before optimization, 40.5--47.3\% of selected-action ratios already fall outside $[0.95,1.05]$, and 19.7--27.2\% of informative choices meet the surrogate-clipping condition. Small average distribution differences can matter for a narrow clipping interval. These measurements do not isolate their contribution to policy drift.

\subsection{Perception Failures Before Learning}\label{app:fault-isolation}
Two distinct failed initial-map images are tested with greedy decoding and four stochastic samples (temperatures 0.2, 0.2, 0.7, 0.7). Neither first-valid nor ensemble recovery succeeds (0/2). Six training-image controls are exact in both conditions, while total decoding time rises from 88.83 to 445.75 seconds. In eight matched short games, failed initial maps stop before action with undefined reward; control games execute eight steps without death.

A separate language-LoRA observation test yields 109 exact valid outputs and eight malformed outputs in 117 observations, with no valid-but-semantically-wrong answer. These cover 32 distinct images, including two producing malformed responses. Collection stops at 125/144 planned attempts in each life setting after eight faults, before any learning update.

\subsection{Statistical Units and Paired Uncertainty}
\label{app:paired-uncertainty}
The observed fraction in a completed panel is a description of those records. Uncertainty intervals require a sampling model. For the action-head comparison we treat the 64 execution-seed pairs within each extra-life setting as independent draws from one fixed-map execution distribution, with both checkpoints fixed. The interval does not cover new layouts, independent training runs, protocol changes or model selection. These assumptions are an inferential idealization of the seed panel, not additional experimental replication.

\paragraph{Preserve the pair.}
Let $Y_0,Y_1\in\{0,1\}$ indicate clearing with three and twenty total updates. Write $p_+=\Pr(Y_0=0,Y_1=1)$ and $p_-=\Pr(Y_0=1,Y_1=0)$. Then $\Delta=p_+-p_-$, estimated by $(n_+-n_-)/64$. We construct separate 97.5\% exact Clopper--Pearson intervals~\citep{clopper1934confidence} $[L_+,U_+]$ and $[L_-,U_-]$ for these two within-pair event probabilities, using 64 as each denominator. The difference bound is
\[
       [L_+-U_-,\;U_+-L_-].
\]
By a union bound, its coverage is at least 95\% under the stated sampling model, without assuming independence between improvements and regressions. It is conservative. It is a per-setting bound, not simultaneous coverage across both extra-life endpoints. This differs from subtracting marginal Wilson bounds as if the two arms were independent.

\begin{table}[ht]
\centering\small\setlength{\tabcolsep}{4pt}
\caption{Paired clearance difference: twenty minus three updates. Units are percentage points; bounds are rounded. Each row uses 64 pairs on one fixed layout.}
\label{tab:paired-uncertainty}
\begin{tabular}{lrr}
\toprule Extra lives & Observed change & Conservative 95\% bound \\
\midrule 0 & $-14.1$ & $[-33.5,+7.3]$ \\
3 & $-7.8$ & $[-29.1,+14.4]$ \\
\bottomrule
\end{tabular}
\end{table}

The 0-extra-life discordances are five improvements and fourteen regressions; 3-extra-life discordances are eight and thirteen. The exact two-sided McNemar test conditions on their totals, giving unadjusted $p=0.0636$ and $p=0.3833$. These are exploratory summaries without a multiple-endpoint correction or a confirmatory model-selection claim. Both conservative bounds include zero. The decline is specific to this fixed-map panel.

\paragraph{Other panels use other units.}
The text-based maze normal-start validations repeat policy samples on eight held-out layouts. A cross-layout interval would resample or model layouts, rather than treating all 96 samples as independent layouts. Perception banks contain temporal frames and repeated decodes, so 106 outputs do not give 106 independent visual scenes. The independent 60-game profile is structured as four execution seeds by three generation seeds per condition on one map. The Codex result remains one preliminary game. We retain their counts and scopes rather than assigning every denominator the same binomial interpretation.

\paragraph{Conditional costs.}
Calls and steps on jointly successful pairs describe a selected subset that depends on both policies. Across all attempts, early deaths change interaction length. Conditional savings therefore do not establish population-wide efficiency or a causal cost advantage.

\section{Step 4: Joint Training and Harness Improvement}
Step 4 combines RL with improvements to visual grounding, state maintenance and candidate execution. The following directions build on the preceding experiments and preliminary observations.
\subsection{Discussion: Deciding What to Fix}\label{expanded:sec:discussion}

The central question is which component to improve. Table~\ref{tab:guide} links perception, state maintenance and decision failures to prospective tests that measure component accuracy and full-game outcomes together.

\begin{table}[t]
\centering
\small
\begin{tabular}{p{0.17\columnwidth}p{0.38\columnwidth}p{0.30\columnwidth}}
\toprule
Stage & Evidence in this study & Next intervention to validate \\
\midrule
Perception & Transcription errors and separate static/closed-loop query evaluations & Align prompt and format; assess visual adaptation under its declared protocol \\
\addlinespace
State & Actor estimates overwrite terrain; typed updates prevent this in short windows & Enforce typed writes and explicit unknowns; compare full episodes \\
\addlinespace
Decision & Same-stage training drops; no gain on the paired action-head panel & Retain matched risk-check comparisons; test corrective supervision separately \\
\bottomrule
\end{tabular}
\caption{From recorded evidence to the next test. Harness comparisons support their named mechanisms; revised memory and corrective supervision still require their own outcome comparisons.}
\label{tab:guide}
\end{table}

\paragraph{Learning signals and control.}
Textual primitive-action learning improves normal-start completion but still fluctuates within stages. Additional output-head training shows no benefit on its fixed-map panel. Matched reward and supervision comparisons can test denser credit assignment, including reward decomposition~\citep{vanseijen2017hybrid}, while holding observation and control interfaces fixed.

\paragraph{Accurate state versus usable motion.}
Correct player, pellet and timer estimates coexist with unknown active-ghost headings in 291 of 431 observations. Replacing missing headings changes local gate acceptance (Appendix~\ref{app:public-memory-followup}), motivating deployable motion estimators and full-game evaluation. Separately, visual language-LoRA training shows policy drift and uninformative reward groups. Motion estimation and reward discrimination therefore require separate tests.

\paragraph{Fit to geometric and spatial reasoning.}
The failures we observe are spatial: localising entities to the right grid cell, maintaining a consistent topology over time, and reasoning about whether a ghost can reach an intersection first. Our results suggest that spatial-reasoning benchmarks for VLMs should evaluate these abilities in the loop and over time, because the static versions can be nearly solved while the closed-loop task is not.

\paragraph{Future work: learning from failures.}
Future researchers could use the student's actual failure states to select corrective exercises for post-training RL, compare a search-based teacher with simpler rules, and test whether action supervision or optional reasoning supervision transfers to independent play. The central evaluation must remove the teacher and extra reasoning assistance, check for regression on prior skills, and account for teacher and training costs. Our small single-goal intervention screen (Appendix~\ref{app:early-exploration}) tests assistance at inference, not learned improvement.

\paragraph{Limitations.}
The study concerns one game and primarily one model family. Experiments with different life budgets, observations, fallback rules and task distributions are reported separately. Full visual-agent evaluation on unseen layouts, repeated training seeds and controlled full-game state-update comparisons remain open. Some early visual training used prefix caching that introduced probability differences; its effect was not isolated. Component substitutions cover local states or short windows rather than complete causal decompositions.

\subsection{Early single-goal intervention screen}\label{app:early-exploration}
On eight ghost-free single-pellet cases (frozen model, 256 actions), direct play, one reminder and short thinking each completed \pend{4/8}; one BFS action followed by student control completed \pend{5/8}. Five cases triggered each intervention; all thinking calls used the 256-token cap. BFS rescued one loop; three direct failures remained. This screen measures inference-time assistance on single-pellet tasks.

\label{app:diagnostic-detail}
\paragraph{Controlled component comparisons.}
A state-update comparison should vary only the update rule while matching initial state, public feedback, policy, candidates and action budget. Initial-map errors should remain visible. Fixed actions measure update accuracy; closed-loop games measure behavioral benefit. Motion estimators should be scored against explicit heading targets and then tested in full games, retaining deaths, faults and inference costs.

\paragraph{Implications for spatial-agent evaluation.}
Separate geometry from decisions, distinguishing walls, consumable pellets and moving actors; cell accuracy hides persistent errors. Declare motion targets: past displacement differs from future heading after a turn. State each score's and comparison's information access for pixels, feedback, rules and private state; engine positions change the screenshot task. Retain outcomes, field accuracy, costs and within-seed pairing; distinguish new topologies from new executions. Fixed-map and held-out text-based maze panels do not replace full visual-agent unseen-layout evaluation. Selected traces do not establish population-level causation.

\section{Visual Primitive-action Learning: Challenges and Settings}\label{app:finding-detail}
\paragraph{Learning difficulties.}
The early visual agent maps each screenshot directly to a direction token and learns with group-relative policy optimization~\citep{shao2024deepseekmath}. A curriculum changes maze size and ghost count. After five to six days, later-stage performance has not converged. The following challenges concern exploration, credit assignment and the observation/action interface; stage-specific compute measurements are unavailable.

\begin{itemize}
    \item \textbf{Limited exploration and shortcut risk.} Training-seed fit without matching held-out performance, weak visual transcription and low policy entropy are consistent with layout-specific shortcuts. They do not localize memorisation to the language component or prove that visual error caused entropy collapse. Sparse terminal rewards can reinforce a narrow set of successful trajectories, consistent with the covariance account of entropy collapse~\citep{cui2025entropy}. We used Clip-Cov to limit this concentration (Appendix~\ref{app:training}), but the absent matched with/without curves prevent a causal claim about how much it preserved exploration.
    \item \textbf{Credit assignment under sparse reward.} The only reward came at the end of a long episode. Training was slow and unstable, with reward sometimes dropping for long stretches. When an episode receives positive relative advantage, recovered mistakes can be reinforced along with useful moves. Denser shaped rewards were not a safe fix: in one run, shaped evaluation reward rose from \pend{48.78} to \pend{126.38} while wins fell from \pend{15/96} to \pend{2/96}, consistent with an objective mismatch, without a matched causal comparison. We therefore kept a terminal win reward and grouped returns per episode rather than per intermediate state (Appendix~\ref{app:training}).
    \item \textbf{Cold start and curriculum calibration.} On hard stages an initial policy may never win, leaving groups with no reward variation and no learning signal. The initial model also oscillated at the start. Starting closer to success, as in Backplay~\citep{resnick2018backplay}, is one possible intervention. Our restart-training approach instead samples states before recorded deaths, including from episodes that eventually win. Its training-start concentration and normal-start validation are described in Sections~\ref{app:ascii-transfer-distribution} and~\ref{app:ascii-resume-extension}; it is not a controlled demonstration of Backplay benefit.
    \item \textbf{A single frame is not a Markov state.} One screenshot does not directly expose ghost velocity (normal pupils in this renderer track Pac-Man, not motion; Appendix~\ref{app:native-rules}), separates overlapping ghosts poorly, and does not show how long a frightened ghost has left before it turns dangerous again. Short of reading the engine's internal state, acting well requires memory across frames. Asking the model to emit one action token per step also forces it to plan implicitly, which makes its behaviour a black box.
    \item \textbf{Cost.} Every step is a model call, and an episode has hundreds of steps, so rollouts are expensive and slow down training further. With options, one model decision covers \pend{1.83} environment steps on average, and the amortised decision pause falls from \pend{169} to \pend{148}\,ms per step even though each assisted decision has a longer total observation-to-action pause. These decision-cost measurements come from a separate 12-game-per-condition profile (Appendix~\ref{app:independent-profile}).
\end{itemize}

Exploration involves perception and shortcut risk; credit assignment and cold start concern the learning signal, while partial observability and cost arise from the interface. Even a perfect perceiver needs history to infer ghost heading, speed and frightened timers; the harness provides this memory. This motivated two interface changes: visual adaptation (Appendix~\ref{app:perception-panels}) and placing memory and candidate generation in a harness so that the model chooses among a few routes instead of hundreds of primitive moves (Appendix~\ref{app:factorial-protocol}). Table~\ref{tab:headline} previews the outcome.

\subsection{Historical visual primitive-policy training}\label{app:training}
\paragraph{Scope.} This section describes historical headline runs; the separate action-head settings appear in Appendix~\ref{app:head-rl-protocol}, and text-based maze restart settings appear in Appendix~\ref{app:fsr-training-details}.\par
\paragraph{Optimisation.} Group-relative policy optimisation~\citep{shao2024deepseekmath} with a terminal win reward (1 for clearing all regular pellets within 512 decisions without losing a life). Training on \pend{$8\times$H800} GPUs; inference on one GPU per model. The group size, learning rate, KL coefficient and per-update rollout budget are unspecified; the settings in Appendix~\ref{app:fsr-training-details} apply only to that named continuation.
The episode-grouping calculation appears in Section~\ref{sec:policy-objective}.
\paragraph{Curriculum.} Tiers grow maze size first and ghost count second: \pend{$13{\times}13$/2 ghosts $\to$ $15{\times}15$/2 $\to$ $15{\times}15$/4 $\to$ $17{\times}17$/4 $\to$ $19{\times}21$/4 $\to$ $21{\times}25$/4} (columns $\times$ rows). A tier is promoted when the training win rate over a three-batch window passes a threshold (\pend{70--80\%}) with at least \pend{384} episodes on \pend{8} layouts, on two consecutive checks; promotion takes effect after the optimiser update. \pend{13/16} rollout groups come from the current tier and \pend{3/16} review easier tiers. Validation never triggers promotion. In this configuration every episode starts from the initial state. This ordinary-start configuration is separate from the failure-state restart continuation in Appendix~\ref{app:fsr-training-details}.
\paragraph{Fit versus generalisation.} The pipeline can fit fixed training seeds (training-seed evaluation \pend{83.33\%}), while the best validation win rate in a separate run was \pend{44.17\%}; these results come from different runs and evaluation populations.
\paragraph{Preventing entropy collapse.} We use Clip-Cov~\citep{cui2025entropy}, which detaches the gradient of a small fraction of tokens whose log-probability and advantage have the highest covariance, restricting those selected token gradients; its effect in these runs is not isolated. Historical token fractions, thresholds and matched with/without entropy curves are unavailable. Raising the upper clipping bound (Clip-Higher) and discarding rollout groups with identical rewards (Dynamic Sampling), both from DAPO~\citep{yu2025dapo}, are alternatives that target the same collapse and cold-start groups with no reward variation, respectively; we did not use them.
\paragraph{Restart-state selection.} Appendix~\ref{app:fsr-training-details} gives the failure-state restart settings; Section~\ref{app:ascii-transfer-distribution} details pre-death sampling and source concentration. Source episodes can eventually succeed; these starts are not uniformly sampled backward from a successful terminal state. A controlled success-versus-distance curve remains unavailable.

\subsection{Training and Sampling Consistency}\label{app:pitfalls}
Policy-gradient learning requires sampled actions and evaluated probabilities to use the same legal support, temperature and multimodal representation. Independent episodes must also remain independent when batches are packed. Violating these conditions changes the effective learning problem, rather than merely slowing computation.

Early configurations showed probability differences associated with prefix-cache reuse and numerical precision. Their isolated effect on learning is unknown. The textual restart recipe disables prefix/radix caching and CUDA graphs; this choice is not evidence that every earlier visual run used the same sampling behavior.

\section{Related Work}
\label{app:related-work}

\subsection{Information access and success measures}
Despite its apparent simplicity, Pac-Man results rely on different information conditions and evaluation objectives. Some systems receive game-provided maze, pellet and ghost states rather than infer the environment from screenshots, as in JevMan's direction-selection interface.\footnote{\url{https://opper.ai/jevman-benchmark/leaderboard}} HRA instead uses object-position representations extracted from images; its maximum-score experiment additionally memorizes successful action sequences that do not generalize~\citep{vanseijen2017hybrid}. Pixel-based RL has also made substantial progress~\citep{mnih2015human}: Agent57 exceeds its human-score reference on all 57 Atari games, including Ms.\ Pac-Man, but high cumulative scores do not establish reliable level completion or unseen-maze generalization~\citep{badia2020agent57}. Increasing maze scale presents an additional learning and control challenge. Information access, assistance, task scale and evaluation objectives therefore matter when interpreting performance.

\subsection{Tasks and evaluation}
\paragraph{Interactive game benchmarks.}
Games expose the consequences of sequences of decisions, rather than isolated answers. BALROG evaluates VLM agents across games and compares visual and textual interfaces~\citep{paglieri2024balrog}. lmgame-Bench studies game control with perception and memory support, including transfer through game-specific RL~\citep{hu2025lmgame}. These benchmarks ask how agents perform across tasks and interfaces. Our Pac-Man study instead follows one development process, using controlled harness comparisons and concrete failure cases to study perception, state maintenance and decisions.

\subsection{Potential problems}
\paragraph{Perception, spatial understanding and visual uncertainty.}
Fine-grained tests reveal difficulties with simple patterns and precise visual localisation~\citep{tong2024eyes,rahmanzadehgervi2024blind}. SpatialVLM improves spatial question answering with spatial training data~\citep{chen2024spatialvlm}, while Thinking in Space examines spatial recall from sequential visual observations~\citep{yang2025thinking}. Uncertainty-aware mapping represents uncertain semantic readings~\citep{bajpai2025uncertainty}; monoVLN acquires further views when observations are incomplete~\citep{lu2025monovln}. These works distinguish unreliable readings from missing information. Our largely visible board can still be misread, and the resulting error can affect later state and actions. Static answer accuracy therefore measures a different requirement from reliable closed-loop control.

\paragraph{State memory and updating.}
Remembered information can become stale, as studied by When Memory Lies~\citep{sun2026memorylies}; EvolvingNav maintains time-indexed beliefs about moving objects outside the current view~\citep{gao2026evolvingnav}. Belief revision provides a classical foundation for reconciling new evidence with retained beliefs~\citep{boutilier1998belief}. Our fixed-maze case concerns a complementary failure: a new actor misreading overwrites correct terrain. Separating actor estimates from static terrain prevents this overwrite in short windows; full-game benefit still requires evaluation. State uncertainty is thus broader than uncertainty in the current image.

\paragraph{Decision, planning and skill execution.}
Even useful observations and retained state must be turned into feasible action sequences. Uncertainty about action consequences can persist after image transcription is removed; it is not simply visual uncertainty. Our textual controls, action-head training outcomes and fixed-weight harness comparisons address different parts of this problem. They provide evidence about particular decision and execution limitations, without completing full-episode causal replacements of perception, memory and policy.

\subsection{Solving approaches}
\paragraph{Teacher supervision and imitation.}
PlayJev is a model release that first imitates programmatic action teachers through behavior cloning, then obtains teacher action labels for states visited by the learner~\citep{omnijev2026playjev}. DAgger (Dataset Aggregation) adds these visited-state labels to earlier data and retrains, addressing distribution shift and compounding errors when the learner's mistakes take it away from demonstration states~\citep{pmlr-v15-ross11a}. This supplies a dense-supervision alternative to our reward-driven action training. We do not assume an additional competent action teacher, while retaining pretrained models and environment-derived perception supervision. Different game and observation conditions prevent treating the releases as a matched performance comparison.

\paragraph{RL and post-training.}
DQN learns game control from reward~\citep{mnih2015human}; hybrid reward decomposition addresses the structure of Ms.\ Pac-Man control~\citep{vanseijen2017hybrid}. VAGEN trains multi-turn visual agents to reason about states and transitions~\citep{wang2025vagen}. These approaches motivate learning better decisions, but a changed overall score does not by itself identify which component improved. We report protocol-specific training outcomes alongside comparisons that hold model weights fixed. Our contribution is evidence-grounded development experience under these protocols.

\paragraph{Curricula and start-state design.}
Curriculum learning organizes practice by difficulty~\citep{bengio2009curriculum}. Backplay moves starts backwards through a successful demonstration~\citep{resnick2018backplay}; reverse curriculum generation expands starts outwards from a goal~\citep{florensa2017reverse}. Both organize practice to make useful learning signals easier to obtain. Our failure-state restart lineage instead samples states before recorded deaths. Concentrated training starts and weaker normal-start validation motivate checking the training distribution, but do not isolate a causal benefit or harm of reverse curricula.

\paragraph{Harnesses, robust updates and executable options.}
NavHarness combines goal generation, completion verification and memory compression~\citep{shi2026navharness}. Robust mapping rejects unreliable constraints~\citep{sunderhauf2012switchable}, and occupancy integration fuses uncertain evidence~\citep{hornung2013octomap}. Skill selection~\citep{ahn2022saycan} and temporally extended options~\citep{sutton1999options} structure decisions around executable behavior. Our fixed-weight comparisons examine bounded risk checking and route binding under a shared candidate scaffold within their named panel. These comparisons show how route availability and execution affect an otherwise fixed policy.

\section{Reproducibility and Disclosure}
\subsection{Available Materials}\label{app:artifact-scope}
The companion site, \url{https://where-it-breaks-review.vercel.app/}, provides the manuscript and game demonstrations. We plan to release task configurations, releasable training/evaluation logic and analysis scripts, subject to employer and other rights-holder approval. Original recipe code may use Apache-2.0 where authorized; third-party dependencies retain their own licenses. Model weights, private trajectories and proprietary infrastructure are outside this commitment.

The appendices specify observation interfaces, model variants, action and life budgets, sample sizes and statistical units. Missing original training data and unspecified settings are noted with the relevant experiments. They limit exact replication and should not be filled using settings from a different experiment.

\subsection{Prompts and Outputs}
Appendix~\ref{app:frozen-perception-prompts} gives perception instructions and valid output examples. Decision outputs select one advertised code, whose route meaning depends on the current menu.

\paragraph{AI assistance.}
AI tools assisted drafting, code/source inspection, analysis and presentation. The authors retain responsibility for claims, citations, materials and submission.

\section{Step 1 Supplementary Results}\label{app:result-audits}
\subsection{Normal-start evaluation}\label{app:result-evaluation-accounting}
Each checkpoint is evaluated on eight fixed held-out layouts with twelve samples per layout. For checkpoint $k$, completion is $\widehat p_k^{\mathrm{normal}}=(8\cdot 12)^{-1}\sum_{m=1}^{8}\sum_{j=1}^{12}Y_{kmj}$, where $Y_{kmj}=1$ for a clear within the life and action budgets. Percentages are $100\widehat p_k^{\mathrm{normal}}$. The same panel supports monitoring, checkpoint selection and reporting. Initial and trained policies share layouts and sampling settings. Incomplete per-sample training-map information limits verification of map separation.

The pretrained checkpoint 0 clears 0/96 on the final stage and 2/96 (2.1\%) on the easiest. T7 becomes the main training stage at update 250, already reaching 57/96 at 249; improvement from 0 therefore includes learning in earlier stages. Checkpoint 335 reaches 78/96.

\paragraph{Cross-stage improvement.}\label{app:earlier-stage-transfer}
From update 249 to 335, T1--T7 completion counts change from $(95,96,80,91,83,79,57)$ to $(96,96,79,96,96,90,78)$, with 96 attempts each. After 250, training focuses on T7 with a small T6 rehearsal share and no T4/T5 samples. T4/T5 gains extend beyond direct rehearsal. T6 improves 79/96 to 90/96 under rehearsal, while T3 loses one clear, so the gain is not uniform across stages. The separate restart-distribution sample below includes T2 rehearsal and is not used to establish this interval's exposure pattern.

Training groups contain twelve episodes. First-pass and reward-filtered rates each summarize 192 collection slots with overlapping samples. Reward filtering selects groups during collection; its contribution to the reported rate is kept separate from completion. Life and speed bonuses distinguish game reward from binary success.

\subsection{Transfer to the next task configuration}
\label{app:ascii-transfer-distribution}

\paragraph{T7-to-T8 transfer.}
Checkpoints 270 and 310 each clear 0/96 normal-start T8 games (Table~\ref{tab:ascii-t8-audit}). Both use eight layouts, twelve samples per layout, four normal ghosts, 192 ordinary pellets, four power pellets, temperature 0.7, three extra lives and 512 actions. The textual observation includes live-state and navigation summaries. Evaluation begins normally without restart, action prefix or outcome filtering.

\begin{table}[!ht]
\centering\small
\begin{tabular}{lrrrr}
\toprule
Checkpoint & Clears & Game over & Time limit & Decisions \\
\midrule
270 & 0/96 & 92 & 4 & 25,066 \\
310 & 0/96 & 84 & 12 & 28,994 \\
\bottomrule
\end{tabular}
\caption{Independent normal-start T8 evaluations on the original eight-map panel: each checkpoint clears 0/12 on every layout. Decision counts are summed over 96 attempts per checkpoint, sampled repeatedly on eight layouts. More time-limit and fewer game-over terminations show neither improved completion nor an isolated restart-training benefit.}
\label{tab:ascii-t8-audit}
\end{table}

T8 changes maze topology and size together with the ordinary-pellet objective (118 to 192). The zero-clear results expose a combined task-shift boundary, not an isolated maze-size or curriculum effect.

In the same original restart lineage, T7 normal-start validation records 77/96 at checkpoint 270, 27/96 at 305 and 61/96 at 310. The independent T8 evaluations above use checkpoints 270 and 310; each stage has its own eight-layout panel.

\paragraph{Distinct restart samples.}
The restart-distribution sample below clears 192/192 at batch 333, whereas the continuation in Figure~\ref{fig:ascii-curve335} clears 180/192 (93.75\%) at its batch 333. The eleven-state concentration statistics apply to the former, separately sampled training run.

\paragraph{Why training wins differ from full-task mastery.}
The 192/192 training sample includes 132 T7 restart attempts, 24 T7 ordinary starts and 36 T2 restart attempts, arranged in sixteen groups of twelve. This sample is already all-win before filtering. Success bonuses can differ, but whether they create relative advantage depends on the chosen all-success-group rule (Appendix~\ref{app:grpo-conventions}).

Of 132 T7 restart attempts, 96 begin with at most 15 of 118 ordinary pellets remaining, including 48 with one pellet. The eleven starts come from two maps and three episodes that eventually clear despite earlier deaths. Restart suffixes average 117.05 actions (range 9--354), versus 362.71 (299--434) for ordinary starts in the same batch. The total action budget is preserved. Normal-start validation at checkpoint 330 clears 54/96, but uses different weights and starts from the training sample, so the two rates do not estimate a matched generalization gap. The next training sample yields 114/120 T7 restart clears, 16/36 T7 ordinary-start clears and 36/36 lower-stage clears.

\paragraph{Source diversity.}
During updates 271--320, the 64-state bank refreshes frequently but draws on only two to five maps and four to twelve source episodes at a time. First-pass clears are 4,081/6,240 for T7 restarts, 986/1,560 for T7 ordinary starts and 1,685/1,800 for lower-stage starts. These task distributions differ and are not treatment/control groups. Frequent replacement of saved states does not ensure broad practice.

\subsection{Training drops and sparse credit}\label{app:training-drop-signal}
Within one T7 restart-training run, batch 299 to 300 first-pass rates decrease from $0.8489583333$ to $0.484375$, while filtered rates change from $0.8489583333$ to $0.546875$. Each has 192 slots: 163 to 93 clears before filtering and 163 to 105 afterward. These are collection-batch statistics, not independent post-update evaluations. Starting states and group composition also vary.

Shared episode advantages supply coarse action credit: equal-return groups give zero signal, while positive advantages can reinforce corrected mistakes. The all-success/all-failure-group rule is not fully specified for these batches (Appendix~\ref{app:grpo-conventions}). Sparse reward is therefore a mechanism to test, not the uniquely identified cause of this drop.

\end{document}